%% file: main.tex
\documentclass{article} 
\usepackage{iclr2026_conference,times}

\input{math_commands.tex}

\usepackage{amsmath, amssymb}

\usepackage{hyperref}
\usepackage{url}
\usepackage{booktabs}
\usepackage{multirow}
\usepackage{graphicx}
\usepackage{tikz}
\usepackage{caption}
\usepackage{subcaption}
\usepackage{wrapfig}
\usepackage{booktabs}
\usepackage{siunitx}
\usepackage{overpic}
\usepackage{algorithm}
\usepackage{algpseudocode}

\newcommand {\ilan}[1]{{\color{purple}[ilan: #1]}}

\usepackage{listings}
\usepackage{xcolor}

\lstdefinelanguage{yaml}{
  keywords={true,false,null,y,n},
  keywordstyle=\color{blue}\bfseries,
  basicstyle=\ttfamily,
  comment=[l]{\#},
  commentstyle=\color{gray}\ttfamily,
  stringstyle=\color{red},
  moredelim=[l][\color{purple}]{:},
  breaklines=true
}

\title{FreeSliders: Training-Free, Modality-Agnostic Concept Sliders for Fine-Grained Diffusion Control in Images, Audio, and Video}

\author{
\begin{tabular*}{\textwidth}{@{\extracolsep{\fill}}cccc@{}} 
\bfseries Rotem Ezra \thanks{Equal contribution} \hspace{0.02em} $^{1}$ & \bfseries Hedi Zisling\footnotemark[1] \hspace{0.02em} $^{1}$ & \bfseries Nimrod Berman\footnotemark[1] \hspace{0.02em} $^{1}$ & \bfseries Ilan Naiman$^{1}$ \\
\bfseries Alexey Gorkor$^{2}$ & \bfseries Liran Nochumsohn$^{1}$ & \bfseries Eliya Nachmani$^{3}$ & \bfseries Omri Azencot$^{1}$
\end{tabular*} \\[0.5em]
\\
$^{1}$Faculty of Computer and Information Science, Ben-Gurion University of the Negev \\
$^{2}$Lightricks \\
$^{3}$School of Electrical and Computer Engineering, Ben-Gurion University of the Negev 
}

\iclrfinalcopy 
\begin{document}

\maketitle

\begin{abstract}
Diffusion models have become state-of-the-art generative models for images, audio, and video, yet enabling \emph{fine-grained controllable generation}, i.e.,  continuously steering specific concepts without disturbing unrelated content, remains challenging. Concept Sliders (CS) offer a promising direction by discovering semantic directions through textual contrasts, but they require per-concept training and architecture-specific fine-tuning (e.g., LoRA), limiting scalability to new modalities. In this work we introduce \emph{FreeSliders}, a simple yet effective approach that is fully \emph{training-free} and \emph{modality-agnostic}, achieved by partially estimating the CS formula during inference. To support modality-agnostic evaluation, we extend the CS benchmark to include both video and audio, establishing the first suite for fine-grained concept generation control with multiple modalities. We further propose three evaluation properties along with new metrics to improve evaluation quality. Finally, we identify an open problem of scale selection and non-linear traversals and introduce a two-stage procedure that automatically detects saturation points and reparameterizes traversal for perceptually uniform, semantically meaningful edits. Extensive experiments demonstrate that our method enables plug-and-play, training-free concept control across modalities, improves over existing baselines, and establishes new tools for principled controllable generation.
An interactive presentation of our benchmark and method is available at: \href{https://azencot-group.github.io/FreeSliders/}{https://azencot-group.github.io/FreeSliders/}.
\end{abstract}

\section{Introduction}
Diffusion models have emerged as state-of-the-art generative models, capable of producing realistic and diverse outputs across images, audio, and video~\citep{rombach2022high, ho2022video, shi2023mvdream}. Beyond generating high-quality samples, a central task is \emph{controllable generation}, the ability to steer the generative process along user-specified signals ~\citep{ liu2023audioldm, ho2022video}. In particular, \textit{text-to-x}, where x is a certain modality, has emerged as a powerful control signal for generative models, offering an intuitive human interface and enabling semantically aligned control ~\citep{zhang2023text, zhang2023survey}. This text-guided capability plays a central role in creative applications, allowing users to produce high-quality content without requiring technical knowledge or professional design skills. An abundance of methods have been proposed to enable flexible, accurate generation and editing of creative content  \citep{gal2023image, chiu2025text, gaintseva2025casteer}. However, relying solely on text prompts makes it difficult to precisely edit specific attributes and \textbf{modulate them continuously} without affecting unrelated components, for example, adjusting a person's age without altering their identity, or changing weather intensity without modifying the scene location. This limitation restricts creators' ability to fully realize their intended specific edits.

To enable \textbf{fine-grained, text-driven control} of specific attributes, \citet{gandikota2024concept} introduced \emph{Concept Sliders} (CS), which derives semantic directions from textual contrasts between opposing concepts to target localized edits (e.g., adjusting a person's apparent age). However, CS \emph{requires per-concept training}, which is impractical in interactive settings where users may request diverse, unpredictable concepts that cannot be pre-trained. Moreover, CS and others such as \citep{sridhar2024prompt,choi2021ilvr} often depend on external classifiers or architecture-specific fine-tuning (e.g., LoRA), which may not be available for new backbones, may become infeasible for future models, or may demand deep, model-specific integration. These constraints hinder adoption in \emph{new modalities} (video, audio) and in rapidly evolving open-source models, motivating \emph{architecture- and modality-agnostic, training-free} methods that retain fine-grained control without per-concept training or architectural modifications.

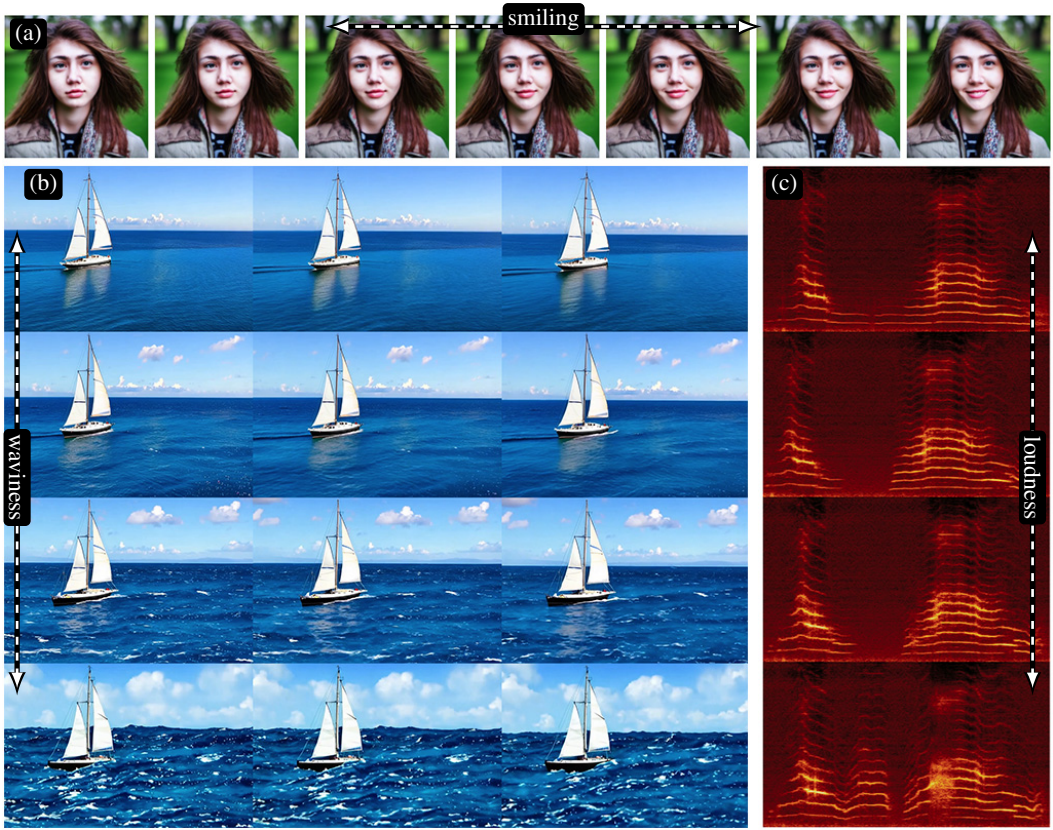
\begin{figure}[t!]
    \centering
    \begin{tikzpicture}
        \node[inner sep=0pt, above right] (img) {\includegraphics[width=\linewidth]{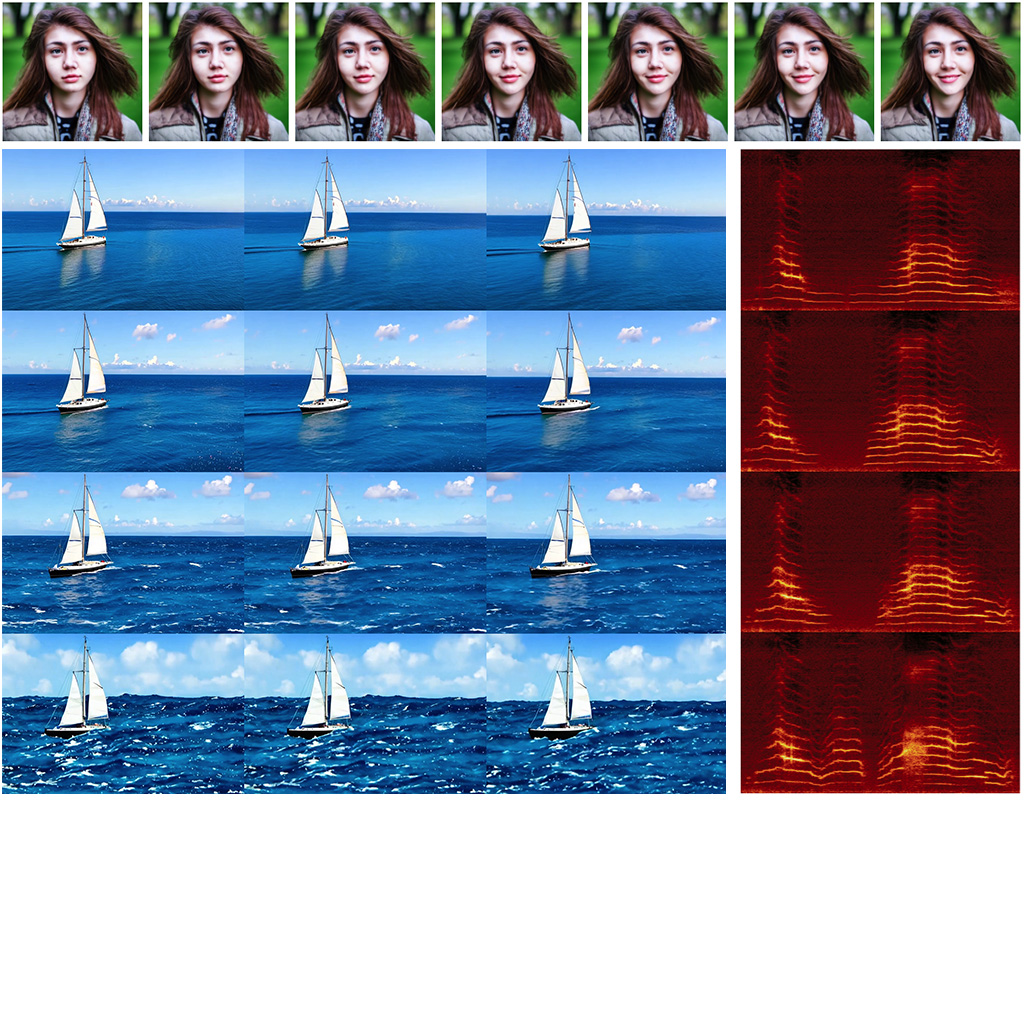}};

        \node[fill=black, text=white, font=\footnotesize, inner sep=2pt, rounded corners=2pt] at (0.35,13.7) {(a)};
        \draw[draw=black, line width=2.0pt, -latex] (6.7,13.8) -- (10.1,13.8);
        \draw[dashed, draw=white, line width=1.0pt, -latex] (6.7,13.8) -- (10.0,13.8);        
        \draw[draw=black, line width=2.0pt, -latex] (6.7,13.8) -- (4.3,13.8);
        \draw[dashed, draw=white, line width=1.0pt, -latex] (6.7,13.8) -- (4.4,13.8);
        \node[fill=black, text=white, font=\footnotesize, inner sep=2pt, rounded corners=2pt] at (7.2,13.9) {smiling};

        \node[fill=black, text=white, font=\footnotesize, inner sep=2pt, rounded corners=2pt] at (0.55,11.7) {(b)};

        \draw[draw=black, line width=2.0pt, -latex] (0.2,8.0) -- (0.2,4.9);
        \draw[dashed, draw=white, line width=1.0pt, -latex] (0.2,8.0) -- (0.2,5.0);
        
        \draw[draw=black, line width=2.0pt, -latex] (0.2,6.0) -- (0.2,11.1);
        \draw[dashed, draw=white, line width=1.0pt, -latex] (0.2,6.0) -- (0.2,11.0);
        \node[fill=black, text=white, font=\footnotesize, inner sep=2pt, rounded corners=2pt]
             at (0.2,7.8) {\rotatebox{-90}{waviness}};

        \node[fill=black, text=white, font=\footnotesize, inner sep=2pt, rounded corners=2pt] at (10.4,11.7) {(c)};
        \draw[draw=black, line width=2.0pt, -latex] (13.7,8.0) -- (13.7,4.9);
        \draw[dashed, draw=white, line width=1.0pt, -latex] (13.7,8.0) -- (13.7,5.0);
        \draw[draw=black, line width=2.0pt, -latex] (13.7,6.0) -- (13.7,11.1);
        \draw[dashed, draw=white, line width=1.0pt, -latex] (13.7,6.0) -- (13.7,11.0);
        \node[fill=black, text=white, font=\footnotesize, inner sep=2pt, rounded corners=2pt]
            at (13.7,7.8) {\rotatebox{-90}{loudness}};
    \end{tikzpicture}

    \vspace{-30mm}
    \caption{\textbf{Sliders:} (a) image: gradually increasing/decreasing smile intensity; (b) video: progressively increasing ocean waviness in a sailing scene; (c) audio: spectrograms of a cat’s meow with rising energy in dominant frequencies from top to bottom, indicating successful concept control.}
\end{figure}

In this work, we propose a simple and elegant CS-inspired solution that is \textbf{training-free and can be seamlessly adapted to new modalities and architectures.} Rather than training a model to directly predict the CS update, we partially estimate the CS formula during inference. This design decouples the method from particular model architecture or modality, thereby enabling plug-and-play concept control method. While our approach is theoretically applicable for any modality, current public benchmarks focus primarily on images. To address this gap, we fundamentally extend the \cite{gandikota2024concept} image benchmark to include standardized \emph{image}, \emph{video} and \emph{audio} concept editing, constituting, to the best of our knowledge, the first modality-agnostic benchmarking for this task. We hope that this benchmark will help drive the development of architecture or modality-agnostic and training-free methods for fine-grained concept manipulation. Finally, on the extended benchmark, our extensive experiments demonstrate that the proposed simple solution \textbf{enables fully training-free concept controlling while achieving improved results}, albeit with a modest increase in computational cost during inference. 

Extending the benchmark to multiple modalities raises non-trivial evaluation questions: How should we measure \emph{fine-grained} control, and can a unified protocol work across modalities? Revisiting work by \citet{gandikota2024concept}, we note that CS evaluations rely largely on $\Delta$CLIP; however, we find that this metric often misaligns with human perception, and its interpretability, i.e., clarifying \emph{why} a slider is 'good', can be vague, especially for users without technical experience. We therefore introduce three \textbf{modality-agnostic properties} that an effective slider should satisfy: \emph{range} (extent of controllable variation), \emph{smoothness} (consistency and monotonicity of intermediate transitions), and \emph{preservation} (extent of non-target content change). To capture these properties, we introduce two new metrics grounded in textual alignment and perceptual similarity.

In our experimentation, we observe a fundamental challenge shared by our method, the original CS approach, and other variants: all require a \emph{scale} that specifies how much to increase or decrease a given property. Two issues arise. (i) It is unclear where a concept \emph{saturates} (e.g., what scale corresponds to the ``maximum age'' of a person). (ii) Because diffusion traversals are inherently non-linear, linearly stepping the scale often yields non-linear semantic progress, which confuses users and hinders intuitive interaction (e.g., skipping age ranges unless many fine steps are used, which is computationally expensive). To resolve this, we introduce \textbf{Automatic Saturation and Traversal Detection (ASTD)}, a two-stage procedure that (1) estimates saturation by combining \emph{conceptual} and \emph{perceptual} scores with a user-tunable trade-off, and (2) fits a data-driven \emph{reparameterization curve} that rescales the traversal so intermediate steps correspond to perceptually uniform, semantically meaningful changes. Empirically, ASTD has been found to improve the overall score (up to $\sim2\times$).

In summary, we present a simple, architecture- and modality-agnostic approach to fine-grained concept control that requires neither per-concept training nor architectural adaptations, enabling seamless use across images, video, and audio. We extend the CS benchmark into the first modality-agnostic evaluation suite for fine-grained control and introduce metrics that more faithfully capture fine-grained control. We further tackle scale selection, saturation, and non-linear traversals with \textbf{ASTD}, a two-stage procedure that detects saturation points and reparameterizes traversals for perceptually uniform changes. Together, these contributions provide a principled, practical, and extensible framework for training-free, fine-grained concept manipulation across modalities.

\section{Related Work}

\paragraph{Controllable generation.} Controllability has long been recognized as a fundamental challenge in generative modeling. Early methods for text-to-image diffusion focused on prompt engineering and classifier guidance, which allow only coarse influence over the output distribution~\citep{dhariwal2021diffusion}. Recent subsequent approaches introduced more explicit control mechanisms, such as attention manipulation~\citep{meng2022sdedit, hertz2023prompt2prompt} and instruction-based fine tuning~\citep{brooks2023instructpix2pix}. Another prominent line of work focuses on personalization, enabling user-specific concepts to be embedded into pre-trained models through techniques such as Textual Inversion~\citep{gal2023image}, DreamBooth \citep{ruiz2023dreambooth}, and LoRA-based fine-tuning \citep{gal2023encoder,tewel2023key} or others \citep{gaintseva2025casteer}. 
While these methods achieve impressive results, they typically require additional training or are tailored to specific modalities or architectures, which limits their scalability and ease of use. These limitations motivate the search for more lightweight and broadly applicable mechanisms of control.

\paragraph{Finding semantic directions.}
A complementary research direction seeks to identify semantic directions in the latent or parameter space of generative models, enabling interpretable traversal along meaningful attributes. 
In the context of GANs, several works demonstrated that linear directions correspond to factors such as pose, age, or expression~\citep{harkonen2020ganspace, jahanian2020on, shen2020interpreting}. 
Recent work extends this paradigm to diffusion models, proposing methods such as Concept Sliders~\citep{gandikota2024concept}, Prompt Sliders~\citep{sridhar2024prompt}, CASteer~\citep{gaintseva2025casteer}, NoiseCLR~\citep{dalva2024noiseclr}, and Concept Steerers~\citep{kim2025concept}. 
A closely related line of work, the disentanglement literature, aims to learn latent factors that separate content, style, and other attributes, often via inductive biases or supervision that promotes factor-wise control~\citep{higgins2017betaVAE, kim2018factorvae, locatello2019challenging, yang2023disdiff, barami2025disentanglement, berman2023multifactor}. Sequential disentanglement further extends this goal to time-varying data, seeking factors that remain stable across time while isolating dynamics (e.g., motion vs.\ identity) for temporally coherent control~\citep{tulyakov2018mocogan, bai2021contrastively, berman2024sequential, naiman2023sample, zisling2025diffsda}. While these methods highlight the promise of semantic directions for fine-grained control, they typically require training additional modules or assume architecture-specific access (e.g., LoRA adapters, text encoders, or cross-attention layers), and have been explored predominantly in the image domain. 
Our work builds on this line of inquiry but departs fundamentally by showing that effective semantic directions can be discovered in a training-free manner and applied consistently across text-to-image, text-to-video, and text-to-audio models.

\section{Architecture-Agnostic, Training-Free Concept Sliders}

In this section, we provide an introduction to diffusion models and the original CS framework. We then present our simple partial inference-time estimation approach.

\subsection{Background}
\paragraph{Diffusion models}\hspace{-3mm}~\citep{ho2020denoising, song2021score} are a class of generative models that synthesize data by learning to reverse a fixed forward noising process. It has proven highly successful across a wide range of modalities, including images, video, audio, and beyond~\citep{luo2021diffusion, rombach2022high, shi2023mvdream, yang2023diffsound, berman2024reviving, po2024state, naiman2024utilizing}. Conceptually, this process begins with a clean data sample, $x_0$, and gradually adds Gaussian noise over a sequence of $T$ timesteps. As $t$ approaches $T$, the data sample $x_T$ is transformed into a sample from a simple prior distribution, typically an isotropic Gaussian, $\mathcal{N}(0, \mathbf{I})$. The generative process learns to reverse this transformation, starting from a random sample $x_T$ and progressively denoising it to produce a realistic sample $x_0$. The core of this reverse process lies in the estimation of the \emph{score function}, which is the gradient of the log-probability of the noisy data $x_t$ with respect to itself. A neural network $s$, parameterized by $\theta$, is trained to approximate this score:
\begin{equation} \label{eq:score_function}
s_\theta(x_t, t, c) \approx \nabla_{x_t} \log p(x_t | c) \ ,
\end{equation}
where $t$ is the timestep and $c$ is conditioning information, such as a text. The score function effectively provides a vector field that points towards regions of higher data density, guiding the generation process. The ability to modify the score function $s_\theta$ during denoising allows for fine-grained control over the generated output, which forms the basis for concept-based methods.

\paragraph{Concept-based methods}\hspace{-3mm} such as Concept Sliders~\citep{gandikota2024concept} and Prompt Sliders~\citep{sridhar2024prompt} manipulate the model's score function to identify \emph{concept directions}. Specifically, Concept Sliders (CS) learn a score function that combines the score at time $t$ with a scaled concept direction, derived from the difference between two conditioning prompts. Formally,
\begin{equation} \label{eq:cs_score}
    \nabla_{x_t} \log p_{\omega}(x_t \mid c_\text{base}) \propto \nabla_{x_t} \log p_{\theta}(x_t \mid c_\text{base}) 
    + \eta \left[ \nabla_{x_t} \log p_{\theta}(x_t \mid c_{+}) - \nabla_{x_t} \log p_{\theta}(x_t \mid c_{-}) \right] \ ,
\end{equation}
where $\eta$ is a user-controlled scaling factor, $c_\text{base}$ is the base conditioning prompt, and $c_{+}$ and $c_{-}$ represent the positive and negative concepts, respectively. The parameters $\theta$ and $\omega$ correspond to the pre-trained diffusion model weights and the trainable LoRA network weights, respectively. Intuitively, this adjustment shifts the generation trajectory toward or away from the target concept while preserving the global structure and identity of the generated sample. In practice, a dataset containing positive, negative, and base examples is prepared for each target concept (e.g., base: ``a car'', positive: ``brand new'', negative: ``wrecked''; the resulting target concept is ``car condition''). Then, the LoRA adapter is fine-tuned to minimize 
\begin{equation}
    \min_\omega \left\lVert \epsilon_\omega(x_t, c_\text{base}, t) - \epsilon_\theta^\text{mod} \right\rVert \ ,
\end{equation}
where $\epsilon_\theta^\text{mod} := \epsilon_{\theta}(x_t, c_\text{base}, t)  + \eta \left[\epsilon_{\theta}(x_t, c_{+}, t) - \epsilon_{\theta}(x_t, c_{-}, t) \right]$ and $\epsilon_\omega, \epsilon_\theta$ are the outputs of the denoising networks. In practice, $\omega$ corresponds to LoRA weights that are fine-tuned per concept, modifying the pre-trained diffusion model weights, i.e., $\theta_\text{new} = \theta + \alpha \omega, \alpha\in\mathbb{R}^{+}$. Controlling the intensity $\eta$ of the target concept can be achieved by modifying $\alpha$ during inference.

As described in the previous sections, current concept sliding methods require a dedicated \emph{training phase} involving:  
(1) collecting datasets for each new concept, and  
(2) fine-tuning a diffusion model with concept-specific parameters.  
While effective, these steps are often infeasible for casual creators who may not have the technical expertise, resources, datasets and time to train models. 

\subsection{Our Approach}

In the original CS framework, the modified score in Eq.~\ref{eq:cs_score} is approximated by training LoRA adapters to match the target modification. At inference, all three required scores  
\[
\nabla_{x_t} \log p_\theta(x_t \mid c_{\text{base}}),\quad 
\nabla_{x_t} \log p_\theta(x_t \mid c_{+}),\quad 
\nabla_{x_t} \log p_\theta(x_t \mid c_{-})
\]
are implicitly encoded in the learned LoRA model. In our \emph{training-free} approach, we compute Eq.~\ref{eq:cs_score} directly from the frozen, pre-trained model without fine-tuning. This is achieved by performing three forward passes per denoising step: one for the base concept, one for the positive concept, and one for the negative concept. Then, the diffusion processes reads:
\begin{equation}
\epsilon(x_t,t) \;=\;
    \begin{cases}
        \epsilon_{\theta}(x_t, c_{\text{base}}, t) \ , & \text{if } t \le k \ , \\ 
        \epsilon_\theta^{\mathrm{mod}}(x_t,c_{\text{base}}, c_+ , c_-, t) \ , & \text{if } t > k \ .
    \end{cases}
\end{equation}

We first run $k$ steps using only the \emph{base concept}. After step $k$, we apply $\epsilon_\theta^{\mathrm{mod}}$ at each remaining timestep. This two-stage process first fixes the \emph{base} concept (the main theme) and then enables fine-grained control over the desired target attribute. Unlike \citet{choi2021ilvr}, we do not use any classifier. Moreover, in contrast to \citet{gandikota2024concept,kim2025concept,sridhar2024prompt}, we do not train additional weights and directly estimate the score. Although our changes are simple, to the best of our knowledge, this setting has not been explored in prior CS-based work. The impact is twofold: (i) by focusing only on the \textbf{sampling}~\citep{karras2022elucidating}, it fully decouples the method from both model architecture and data modality, and (ii) at the cost of modest extra inference compute, it removes the need for any training. We analyze its inference-time overhead and complexity in Sec.~\ref{sec:additional_expreiments}, highlighting the trade-offs. See App.~\ref{sec:psuedo_code} for more details about our inference algorithm procedure.

\section{Evaluation of Concept Sliders}

In this section, we first address two guiding questions: \emph{How do we measure fine-grained control?} and \emph{how can we establish a unified protocol across modalities?} We then revisit current evaluation practices and, finally, propose metric and protocol refinements to close the gaps we identify. 

\subsection{Properties of Slider Evaluation} 
First, we align terminology: a \emph{slider} is any method that, given a target concept, enables explicit control over it. While the motivation for fine-grained control is well established, what constitutes a \emph{good} slider is less explored. Before measuring quality, we propose clear, modality-agnostic properties that serve as objective evaluation criteria:

\paragraph{Range.}  The range of a slider refers to the extent of concept variation it can achieve within a given model.  
For example, for an ``age'' concept, one sliding method might produce outputs spanning from young children to individuals over 100 years old, while another, using the same underlying model, may only reach appearances resembling 70–80 years old.

\paragraph{Smoothness.} The Smoothness of a slider describes its ability to generate \emph{intermediate} conceptual instances in a smooth and perceptually meaningful way.  
In the age example, a slider with a wide range (e.g., 20–100 years old) might still fail to produce realistic intermediate stages such as 50 or 60 years old.  
Thus, a model can have a good range without necessarily being smooth.

\paragraph{Preservation.} Preservation refers to a slider’s ability to maintain the original content and identity of the input while modifying only the target concept.  
For example, when adjusting a person’s age, we aim to retain their original facial identity rather than producing an entirely different individual. This property can sometimes impose a trade-off: certain concepts inherently require altering or discarding aspects of the original identity (e.g., transforming a human into a demon). 

\subsection{Limitations of Current Evaluation}

\newcommand{\metricbox}[2]{%
  {\begingroup\setlength{\fboxsep}{2pt}\colorbox{#1}{\textcolor{white}{\scriptsize #2}}\endgroup}%
}

\begin{figure}[b!]
\centering
\captionsetup[subfigure]{aboveskip=2pt, belowskip=2pt}

\begin{subfigure}{0.48\linewidth}
  \centering
\begin{minipage}{0.08\linewidth}
    \metricbox{black}{2.2}\\[2pt]
    \metricbox{blue!70!black}{4.6}\\[2pt]
    \metricbox{red!70!black}{.28}
  \end{minipage}
  \begin{minipage}{0.90\linewidth}
    \includegraphics[width=\linewidth]{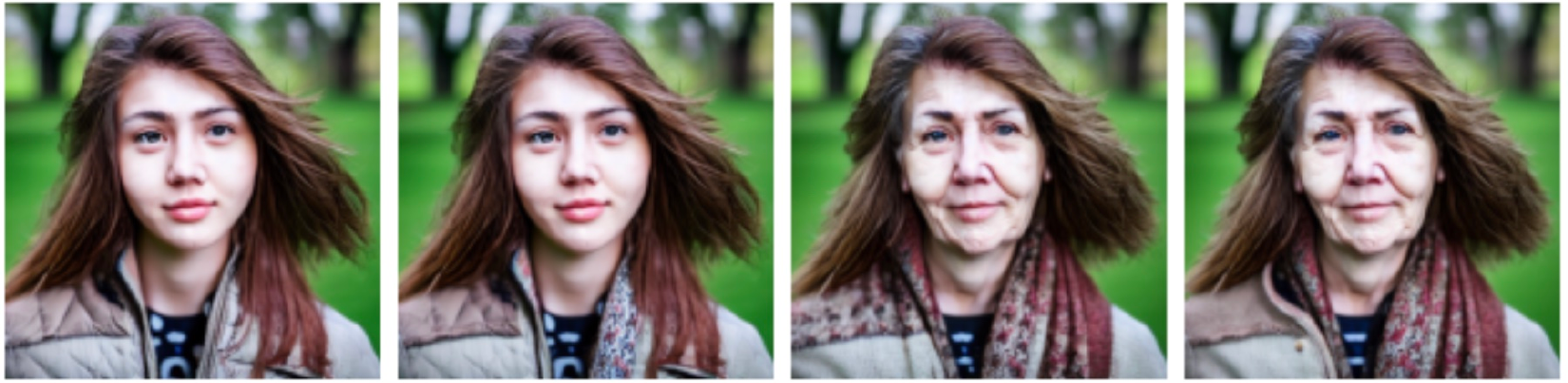}
  \end{minipage}%
\end{subfigure}\hfill
\begin{subfigure}{0.48\linewidth}
  \centering
    \begin{minipage}{0.08\linewidth}
    \metricbox{black}{1.9}\\[2pt]
    \metricbox{blue!70!black}{5.0}\\[2pt]
    \metricbox{red!70!black}{.27}
  \end{minipage}
  \begin{minipage}{0.90\linewidth}
    \includegraphics[width=\linewidth]{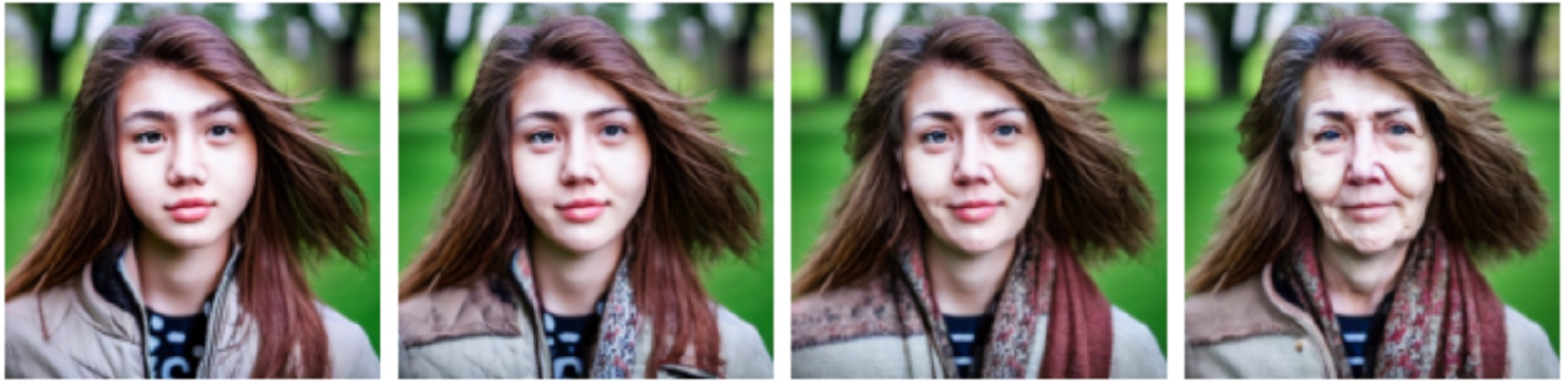}
  \end{minipage}%
\end{subfigure}

\begin{subfigure}{0.48\linewidth}
  \centering
    \begin{minipage}{0.08\linewidth}
    \metricbox{black}{3.9}\\[2pt]
    \metricbox{blue!70!black}{2.8}\\[2pt]
    \metricbox{red!70!black}{.27}
  \end{minipage}
  \begin{minipage}{0.90\linewidth}
    \includegraphics[width=\linewidth]{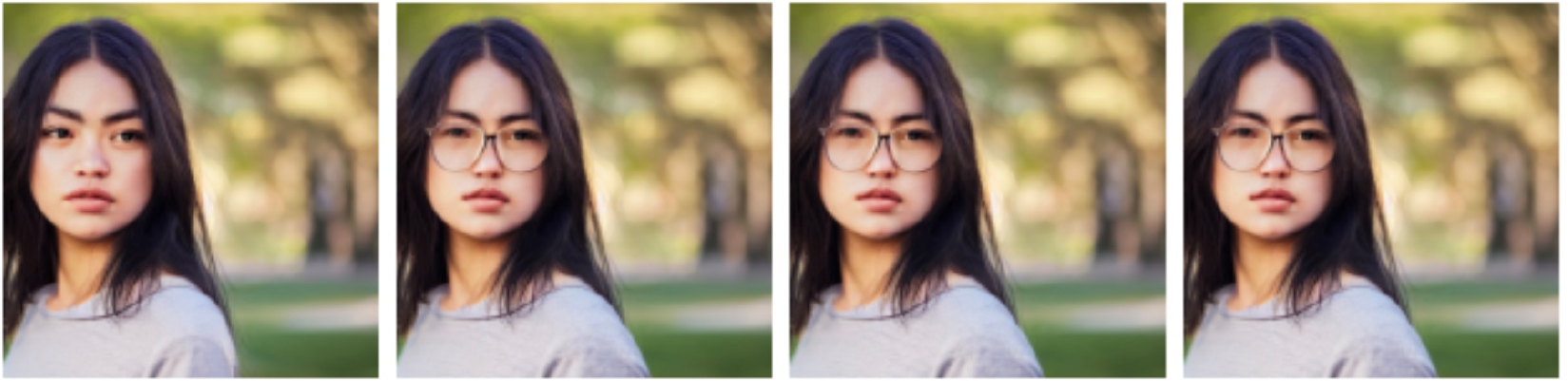}
  \end{minipage}%

\end{subfigure}\hfill
\begin{subfigure}{0.48\linewidth}
  \centering
  \begin{minipage}{0.08\linewidth}
    \metricbox{black}{2.8}\\[2pt]
    \metricbox{blue!70!black}{2.8}\\[2pt]
    \metricbox{red!70!black}{.25}
  \end{minipage}
  \begin{minipage}{0.90\linewidth}
    \includegraphics[width=\linewidth]{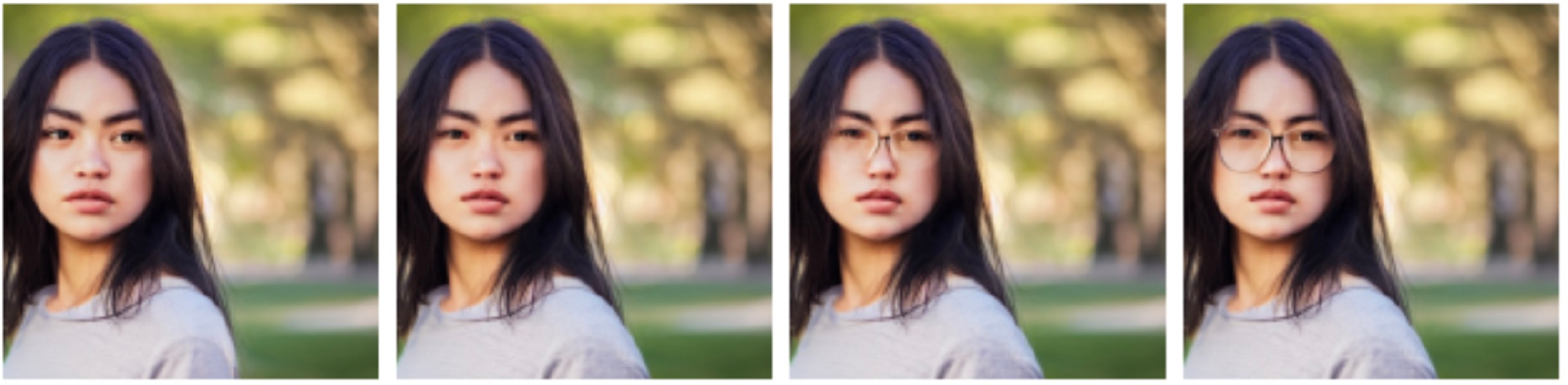}
  \end{minipage}%

\end{subfigure}

\begin{subfigure}{0.48\linewidth}
  \centering
  \begin{minipage}{0.08\linewidth}
    \metricbox{black}{1.1}\\[2pt]
    \metricbox{blue!70!black}{0.6}\\[2pt]
    \metricbox{red!70!black}{.29}
  \end{minipage}
  \begin{minipage}{0.90\linewidth}
    \includegraphics[width=\linewidth]{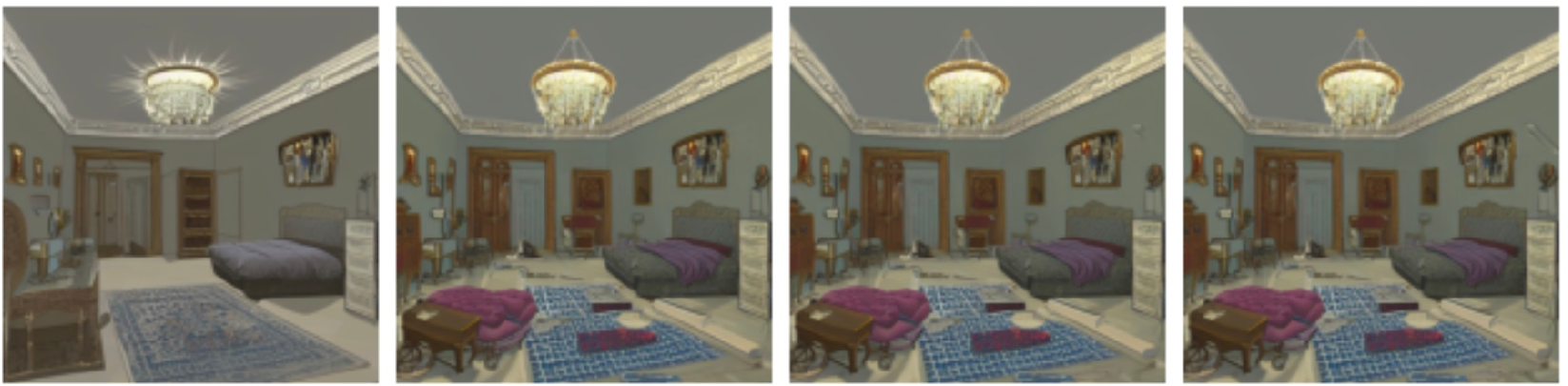}
  \end{minipage}%

\end{subfigure}\hfill
\begin{subfigure}{0.48\linewidth}
  \centering
  \begin{minipage}{0.08\linewidth}
    \metricbox{black}{0.6}\\[2pt]
    \metricbox{blue!70!black}{0.8}\\[2pt]
    \metricbox{red!70!black}{.28}
  \end{minipage}
  \begin{minipage}{0.90\linewidth}
    \includegraphics[width=\linewidth]{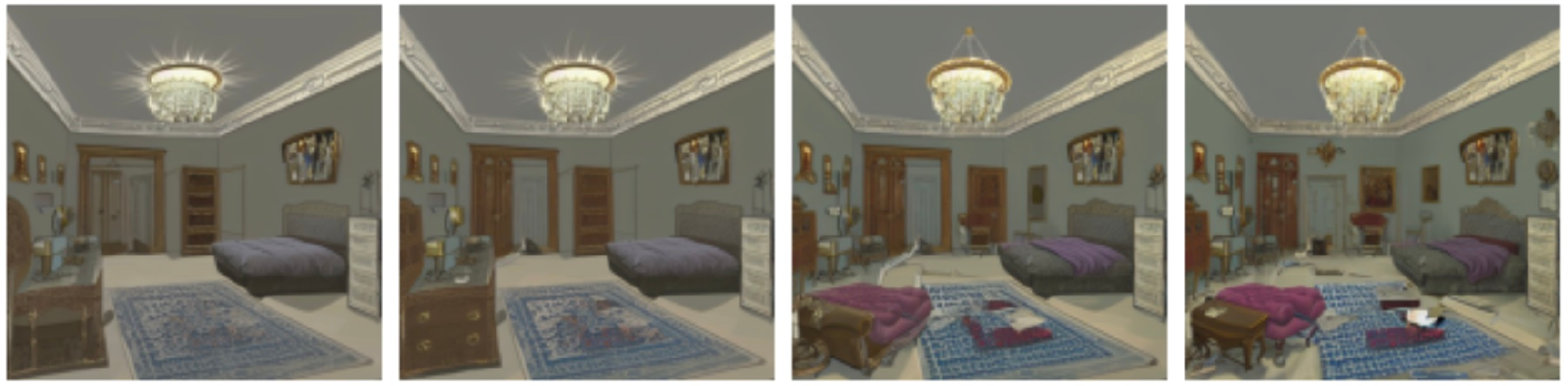}
  \end{minipage}%

\end{subfigure}

\caption{Examples illustrating limitations of the $\Delta$CLIP metric (black boxes): a high score despite abrupt transitions and flat regions at the slider ends (left), and a lower score despite a clear and smooth concept change across intervals (right). Blue and red boxes denote the CR and CSM metrics, respectively, introduced in Sec.~\ref{sec:sliders_metrics}.}
\label{fig:metric-failure-cases}
\end{figure}

Revisiting the standard evaluation protocol of \cite{gandikota2024concept}, two key metrics are used: (1) \textbf{LPIPS} \citep{zhang2018unreasonable}, which measures perceptual similarity between the original and edited images; and (2) \textbf{$\Delta$CLIP}, which quantifies the change in CLIP \citep{radford2021learning} similarity between the original and edited images with respect to a target text prompt, thereby estimating alignment with the desired concept. Under our taxonomy, LPIPS naturally serves as a validator of the \emph{preservation} property and is widely adopted in other benchmarks for the same purposes. Therefore, we retain it as our preservation measure. However, closer inspection of \mbox{$\Delta$CLIP} (see App.~\ref{sec:delta_clip} for full formulation) alongside visual results reveals it often overlooks aspects of \emph{range} and \emph{smoothness}, and in some situations may give misleading scores. This limitation stems from the fact that \mbox{$\Delta$CLIP} was not designed to capture smoothness or ranges of opposite directions. 

In Fig.~\ref{fig:metric-failure-cases}, we compare two sliders side by side. The left column attains a higher $\Delta$CLIP (black box; higher is better) than the right, yet the right column produces a qualitatively better slider. Rows show edits to \emph{age} (top), \emph{glasses} (middle), and \emph{cluttered room} (bottom). These examples reveal that $\Delta$CLIP misses both \emph{range} (extent of change) and \emph{smoothness} (consistency of transitions): in several cases, the visuals remain nearly unchanged or shift abruptly (see App.~\ref{sec:failure_cases_cont} for details).



\subsection{Quantifying Slider Properties with New Metrics}
\label{sec:sliders_metrics}

Building on the previous section, we introduce metrics that alleviate the earlier limitations and explicitly measure the three desired properties. We summarize them here; full definitions and a comparison to $\Delta$CLIP appear in App.~\ref{sec:app_metrics}, ~\ref{sec:delta_clip}.

\paragraph{Semantic Preservation (SP) $\downarrow$:}  
Let $x_\eta$ denote the sample generated with slider scale $\eta$, and let $x_0$ be the original (neutral) sample. Given a set of scales $G$ (excluding $\eta=0$), we define the preservation score as $P(G) = \tfrac{1}{|G|}\sum_{\eta \in G} d(x_\eta, x_0)$, where $d$ is a perceptual distance measure, such as LPIPS for images, LPAPS~\citep{SpecVQGAN_Iashin_2021, paissan2023audio} for audio, or Video-LPIPS~\citep{voleti2022mcvd} for video. This measures the average perceptual deviation of the slider outputs from the original instance. Lower values indicate better preservation of non-target content. This metric complements the others: a trivial method that makes no changes at all will achieve perfect preservation, but very poor range and smoothness. This perceptual measure has been used in the standard CS benchmark.

\paragraph{Conceptual Range (CR) $\uparrow$:}  
To measure how well the slider stretches a concept space, we measure the alignment scores at the endpoints. Let $a(x, c)$ denote the alignment between a generated sample $x$ and a concept prompt $c$. Define: $\text{CR} = 0.5 (\text{CR}_\text{pos} + \text{CR}_\text{neg})$, where
\[
\text{CR}_{\text{pos}} = a(x_{\eta_{\max}}, c_{+}) - a(x_{\eta_{\min}}, c_{+}) \ , 
\quad 
\text{CR}_{\text{neg}} = a(x_{\eta_{\min}}, c_{-}) - a(x_{\eta_{\max}}, c_{-}) \ .
\]
Here $c_{+}$ and $c_{-}$ are the positive and negative prompts, and $\eta_{\min}, \eta_{\max} \in G$ are the minimum and maximum scales. Alignment models $a$ include CLIP for images, ViCLIP~\citep{wang2023internvid} for video, and CLAP~\citep{wu2023latent, htsatke2022} for audio. This metric quantifies the extent of the semantic change. As illustrated in Fig.~\ref{fig:metric-failure-cases}, larger concept shifts yield higher CR. Additionally, unlike $\Delta$CLIP, which measures change only between a neutral point and a single target scale, our metric captures the range property by leveraging both positive and negative directions to represent the full span of variation.

\paragraph{Conceptual Smoothness (CSM) $\downarrow$:}  
Finally, we assess how uniformly the slider distributes concept changes across its scale values. Rather than considering raw alignment scores, we examine the distribution of score increments. For each $\eta \in G$, let $A = \{\, a(x_\eta, c_r) \;\mid\; \eta \in G, \; r = \text{sign}(\eta) \,\}$. We normalize $A$ to the interval $[0,1]$ by rescaling to the maximum attainable alignment score for the concept and assuming the minimum score is 0. 
We split $A$ into its negative and positive subsets and order each subset separately along the slider scale. For each ordered subset, we compute consecutive gaps $ g_i = A_{i+1} - A_i, i=1,\dots, |A|-1$.
and define $\text{CSM} = \mathrm{std}\bigl(\{g_i\}\bigr)$. A lower CSM indicates a smoother and uniform traversal of the concept space; higher values reflect irregular jumps or skipped semantic regions. Thus, a smooth slider maps successive steps to perceptually coherent, evenly spaced changes. As shown in Fig.~\ref{fig:metric-failure-cases}, smoother transitions yield lower (better) CSM.

\section{Saturation and Traversal Detection}

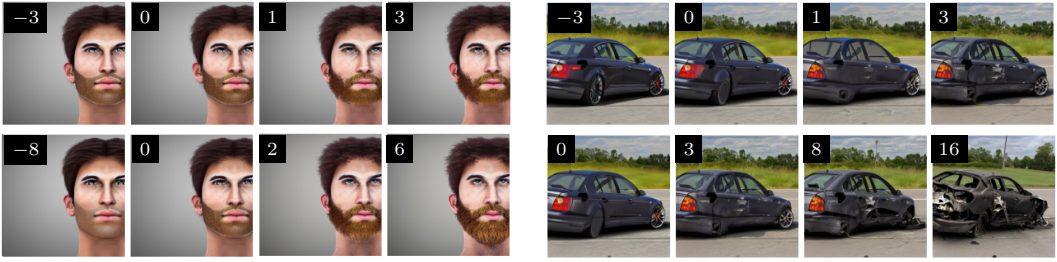
\begin{figure}[t!]
\centering
\makebox[\linewidth][c]{%
  \begin{minipage}[t]{0.485\linewidth}
    \centering

    \begin{overpic}[width=\linewidth]{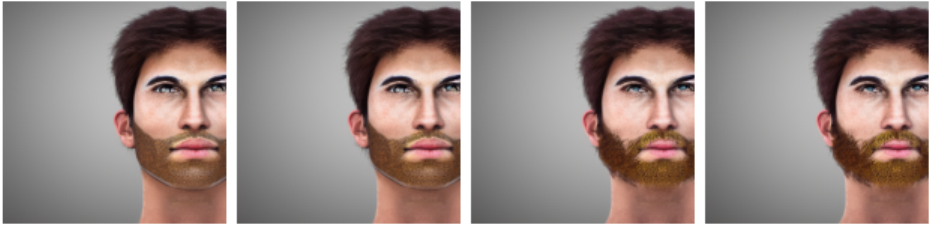}
      \put(0.5,20.5){\colorbox{black}{\textcolor{white}{\scriptsize $-3$}}}
      \put(25.5,20.5){\colorbox{black}{\textcolor{white}{\scriptsize $0$}}}
      \put(50.5,20.5){\colorbox{black}{\textcolor{white}{\scriptsize $1$}}}
      \put(75.5,20.5){\colorbox{black}{\textcolor{white}{\scriptsize $3$}}}
    \end{overpic}

    \vspace{1mm}

    \begin{overpic}[width=\linewidth]{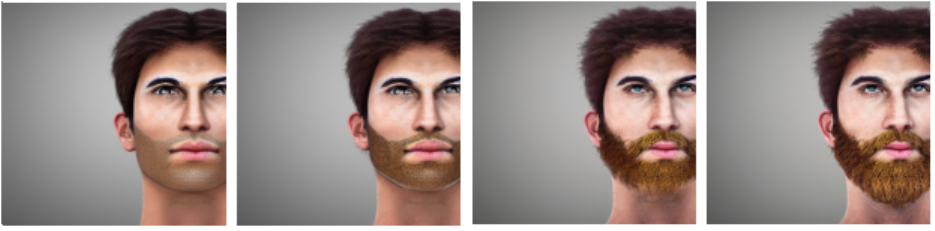}
      \put(0.5,20.5){\colorbox{black}{\textcolor{white}{\scriptsize $-8$}}}
      \put(25.5,20.5){\colorbox{black}{\textcolor{white}{\scriptsize $0$}}}
      \put(50.5,20.5){\colorbox{black}{\textcolor{white}{\scriptsize $2$}}}
      \put(75.5,20.5){\colorbox{black}{\textcolor{white}{\scriptsize $6$}}}
    \end{overpic}
  \end{minipage}
  \hspace{0.02\linewidth}
  \begin{minipage}[t]{0.485\linewidth}
    \centering

    \begin{overpic}[width=\linewidth]{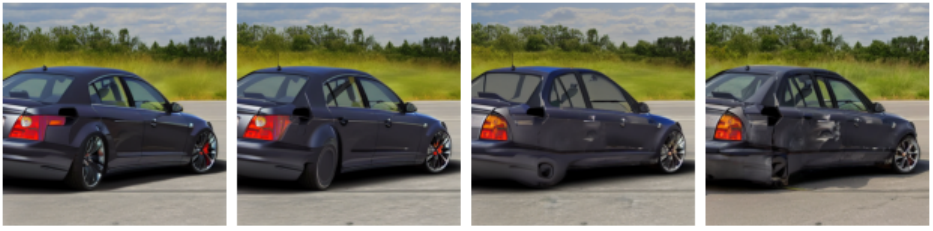}
      \put(0.5,20.5){\colorbox{black}{\textcolor{white}{\scriptsize $-3$}}}
      \put(25.5,20.5){\colorbox{black}{\textcolor{white}{\scriptsize $0$}}}
      \put(50.5,20.5){\colorbox{black}{\textcolor{white}{\scriptsize $1$}}}
      \put(75.5,20.5){\colorbox{black}{\textcolor{white}{\scriptsize $3$}}}
    \end{overpic}

    \vspace{1mm}

    \begin{overpic}[width=\linewidth]{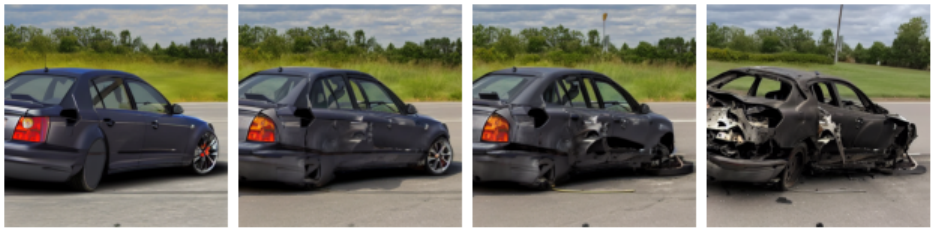}
      \put(0.5,20.5){\colorbox{black}{\textcolor{white}{\scriptsize $0$}}}
      \put(25.5,20.5){\colorbox{black}{\textcolor{white}{\scriptsize $3$}}}
      \put(50.5,20.5){\colorbox{black}{\textcolor{white}{\scriptsize $8$}}}
      \put(75.5,20.5){\colorbox{black}{\textcolor{white}{\scriptsize $16$}}}
    \end{overpic}
  \end{minipage}
}
\caption{Default CS scales often yield suboptimal results as shown in the top row, while the bottom row shows ASTD-optimized scales and step sizes. Black boxes mark sampled $\eta$ values.}
\label{fig:failures-astd}
\end{figure}

The range and smoothness of a slider depend heavily on adjusting the scaling factor $\eta$ for each concept. In prior work, $\eta$ values and intermediate steps are fixed and uniformly distributed across concepts. This leads to two key issues (Fig.~\ref{fig:failures-astd}):  
(1) different concepts operate effectively at different $\eta$ scales, and  
(2) linear traversal in $\eta$ often induces \emph{nonlinear} conceptual progress, resulting in non-smooth changes.  
We therefore propose a two-step procedure to automatically improve slider quality.

\begin{figure}[htbp!]
    \centering
    \includegraphics[width=1\linewidth]{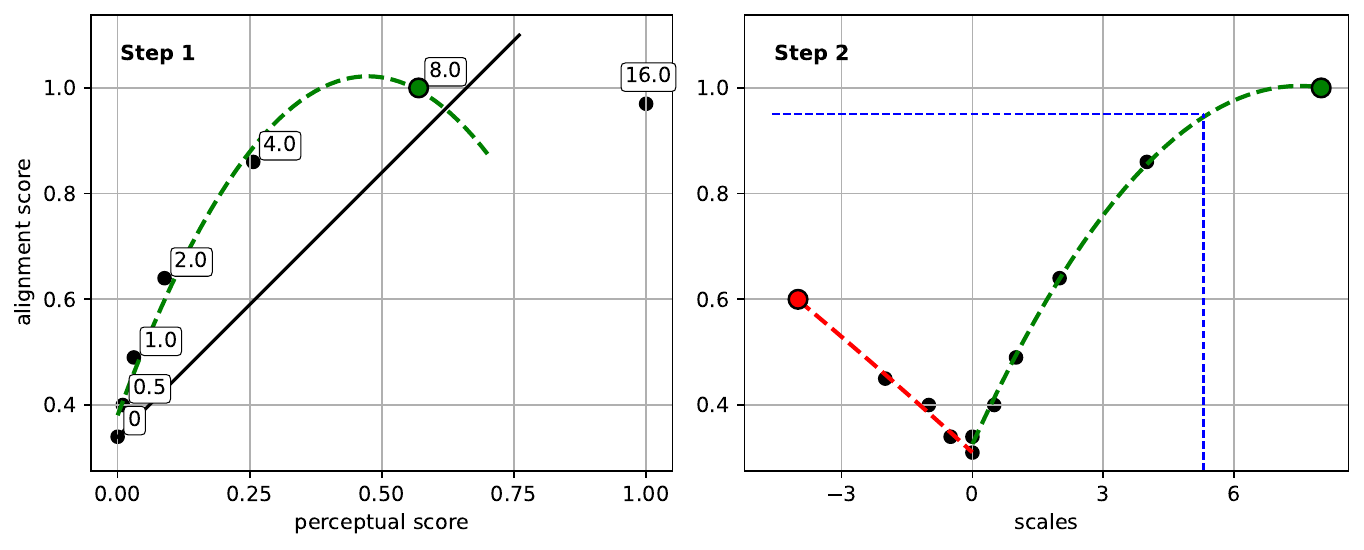}
    \caption{Automatic Saturation and Traversal Detection (ASTD) for concept "smiling". \textbf{Left:} saturation detection via $r(x,\eta)$. \textbf{Right:} traversal reparameterization for smoother progression.}
    \label{fig:astd}
\end{figure}

\paragraph{Step 1: Saturation Detection.}
Determining when a concept saturates is non-trivial.  
Let $a(x,c)$ denote an \emph{alignment score} (e.g., CLIP) measuring the presence or intensity of a concept in a generated image. A naïve approach would declare saturation when $a(x_{\eta},c)$ no longer increases with $\eta$. However, under classifier-free guidance, degraded images can sometimes yield high scores while losing perceptual meaning. To counter this, let $d(x_0,x_{\eta})$ be a similarity measure to the base image (e.g., LPIPS). This ensures edits do not deviate too far from the original content. Since some concepts inherently require larger perceptual changes, we define a user-adjustable trade-off:
\[
r(x,\eta) \;=\; \frac{a(x_{\eta},c)}{d(x_0,x_{\eta})} \ .
\]
The ratio $r$ balances preservation ($r<1$) and concept intensity ($r>1$). In practice, we set $r=1$ (Fig.~\ref{fig:astd}, left, black line). We then evaluate $r(x,\eta)$ at $\eta \in \{0, 0.5, 1, 2, 4, 8, 16\}$ and choose the largest $\eta$ above the reference line as the saturation point (e.g., $\eta=8$ in the figure). The same process is applied to negative scales. Importantly, exploring larger ranges is often necessary: without saturation detection, the original CS (restricted to $[-3,3]$ during evaluation) may entirely miss the effective editing range (Fig.~\ref{fig:failures-astd}). While one could naïvely sweep many scales, this is computationally prohibitive and not user-friendly, motivating our principled approach.

\paragraph{Step 2: Traversal Adjustment.}
After identifying the saturation point, we analyze the curve $\eta \mapsto a(x,\eta)$ to improve traversal smoothness. As shown in Fig.~\ref{fig:astd}, conceptual progression is better captured by a low-degree polynomial than by uniform $\eta$ sampling. We therefore fit a monotone reparameterization that maps the alignment axis to the scale axis, resampling $\eta$ so that equal steps correspond to perceptually uniform changes between $0$ and the saturation point. This adjustment yields smoother, more coherent transitions along the concept axis (Fig.~\ref{fig:failures-astd}), enhancing usability for both the original CS framework and our method.

\section{Results}
This section has two parts. First, we present standardized, independent benchmarks for images, video, and audio. Second, we report additional experiments: (1) adaptability to different backbone models; (2) validation of ASTD’s effectiveness; (3) computational analysis of our method and ASTD; and (4) demonstration of compositional control. Further implementation details for baselines, prompts, concepts, data, and metrics are given in App.~\ref{sec:exp_details}.

\subsection{Multi-Modal Quantitative Benchmark}
\label{sec:image_cs}
We standardize the \emph{Image} benchmark from \citet{gandikota2024concept} and extend it to \emph{Video} and \emph{Audio}. Each modality covers \textbf{ten} concepts. For each concept, we generate 100 sliders in the image benchmark and 10 sliders in the video and audio benchmarks, evaluating each at seven scale values selected as either the default or the ASTD-optimized option, whichever performs better. Importantly, all methods achieve higher overall scores with ASTD. All methods use the same backbone per modality. We first compute the per-concept average across sliders, then report the mean across concepts as the final metric. We use the metrics from Sec.~\ref{sec:sliders_metrics}. For compact reporting, we also provide a single \emph{Overall Score} (OS), $\mathrm{OS} \;=\; \frac{\mathrm{CR}}{\epsilon + \mathrm{SP}} \;+\; (1-\mathrm{CSM})$,

where $\mathrm{CR}$ is \emph{Conceptual Range}, $\mathrm{SP}$ is \emph{Preservation}, and $\mathrm{CSM}$ is \emph{Smoothness}; $\epsilon=1$ stabilizes the ratio for small $\mathrm{SP}$. This design captures the trade-off between conceptual alignment (numerator) and semantic preservation (denominator), augmented by smoothness over the traversal. Qualitative comparisons appear in App.~\ref{sec:qual_comparisons}.

\paragraph{Evaluating image concepts.}
We use \emph{Stable Diffusion v1.4} as the backbone and compare against strong modality-agnostic and/or training-free baselines~\citep{gandikota2024concept,sridhar2024prompt,hertz2023prompt2prompt}. Our method is \emph{training-free}, \emph{modality-agnostic}, and \emph{architecture-independent}. Thus, its important to note, that in those aspects, it has inherent advantages upon the other methods. We also implement a Text Embedding (TE) baseline that applies semantic contrast in text-embedding space rather than score space (see App.~\ref{sec:text_emb}). In Tab.~\ref{tab:bench_results} (Image), our method achieves the largest \emph{CR}, the lowest \emph{SP}, competitive \emph{CSM}, and thus the best overall score.

\paragraph{Evaluating video concepts.}
We use \emph{CogVideoX-2B}~\citep{yang2024cogvideox}. Videos contain both static and dynamic elements: e.g., for \emph{river flow strength}, water motion should vary while background remains stable; for \emph{car style}, dynamics should be preserved while appearance changes. We therefore report both \emph{Dynamic} and \emph{Static} metrics (implementation details in the appendix); table markers \textbf{(D)} and \textbf{(S)} denote these subsets. Final scores follow the image setup. We compare against adapted video versions of CS and Prompt Sliders; Prompt-to-Prompt was not feasible due to CogVideoX’s special attention mechanism. In Tab.~\ref{tab:bench_results} (Video), our method attains higher OS and is competitive or superior on each metric for both static and dynamic evaluations.

\paragraph{Evaluating audio concepts.}
We use \emph{Stable Audio Open}~\citep{EvansPCZTP25}. Final scores follow the same protocol, and we compare against CS and Prompt-to-Prompt. In Tab.~\ref{tab:bench_results} (Audio), our method attains better \emph{CSM} and \emph{SP}, competitive \emph{CR}, and the best OS among all methods.

\begin{table}[tbp]
    \centering
    \caption{Benchmarking results across image, video, and audio concepts, in comparison to several strong baselines. Unavailable entries are shown as “--”.}
    \label{tab:bench_results}
    \resizebox{\columnwidth}{!}{%
    \begin{tabular}{@{}llccc|cc@{}}
    \toprule
    & \textbf{Metric} & \textbf{CS} & \textbf{Prompt Sliders} & \textbf{Prompt-to-Prompt} & \textbf{TE} & \textbf{Ours} \\
    \midrule
    \multirow{4}{*}{\rotatebox{90}{Image}}
     & CR $\uparrow$              & $2.54 \pm .776$ & $.079 \pm .374$ & $.566 \pm .554$ & $.927 \pm .803$ & \underline{$2.85 \pm .861$} \\
     & CSM $\downarrow$       & $.276 \pm .009$ & $.292 \pm .014$ & $\underline{.257 \pm .013}$ & $.285 \pm .010$ & $.283 \pm .009$ \\
     & SP $\downarrow$     & $.062 \pm .031$ & $.065 \pm .032$ & $.125 \pm .029$ & $.019 \pm .013$ & \underline{$.018 \pm .015$} \\
     & Overall Score $\uparrow$      & 3.11 & .782 & 1.24 & 1.62 & \textbf{3.56} \\
    \midrule
    \multirow{7}{*}{\rotatebox{90}{Video}}
     & CR (S) $\uparrow$              & $1.17 \pm .44$  & $-.173 \pm .59$ & --              & $1.30 \pm .55$   & \underline{$1.52 \pm .60$} \\
     & CR (D) $\uparrow$              & $.013 \pm .01$  & $-.001 \pm .01$ & --              & $.021 \pm .01$   & \underline{$.024 \pm .01$} \\
     & CSM (S) $\downarrow$            & $.308 \pm .01$ & $.328 \pm .02$ & --              & $\underline{.306 \pm .01}$  & $.311 \pm .01$ \\
     & CSM (D) $\downarrow$            & $.460 \pm .01$  & $.457 \pm .02$  & --              & $.442 \pm .01$   & \underline{$.436 \pm .01$} \\
     & SP (S) $\downarrow$            & $.761 \pm .02$  & $.112 \pm .05$  & --              & $.041 \pm .01$   & \underline{$.032 \pm .01$} \\
     & SP (D) $\uparrow$              & $-1.56 \pm .94$ & \underline{$-.931 \pm .84$} & -- & $-1.19 \pm .67$ & $-.996 \pm .54$ \\
     & Overall Score $\uparrow$      & 2.29 & 1.04 & -- & 2.40 & \textbf{2.70} \\
    \midrule
    \multirow{4}{*}{\rotatebox{90}{Audio}}
     & CR $\uparrow$                  & $5.78 \pm 1.42$ & --              & \underline{$6.75 \pm 2.23$} & $6.13 \pm 1.72$ & $6.04 \pm 2.04$ \\
     & CSM $\downarrow$                & $.351 \pm .016$ & --              & $.386 \pm .026$ & $.393 \pm .018$ & \underline{$.344 \pm .024$} \\
     & SP $\downarrow$                & $2.72 \pm 0.32$ & --              & $4.19 \pm 0.59$ & $4.21 \pm 0.32$ & \underline{$2.13 \pm 0.42$} \\
     & Overall Score $\uparrow$      & 2.20 & -- & 1.91 & 1.78 & \textbf{2.58} \\
    \bottomrule
    \end{tabular}%
    }
\end{table}

\setlength{\tabcolsep}{2pt}
\subsection{Additional Experiments}
\label{sec:additional_expreiments}

\begin{figure}[t]
\centering
\begin{subfigure}{0.452\linewidth}
    \centering
    \begin{overpic}[width=\linewidth]{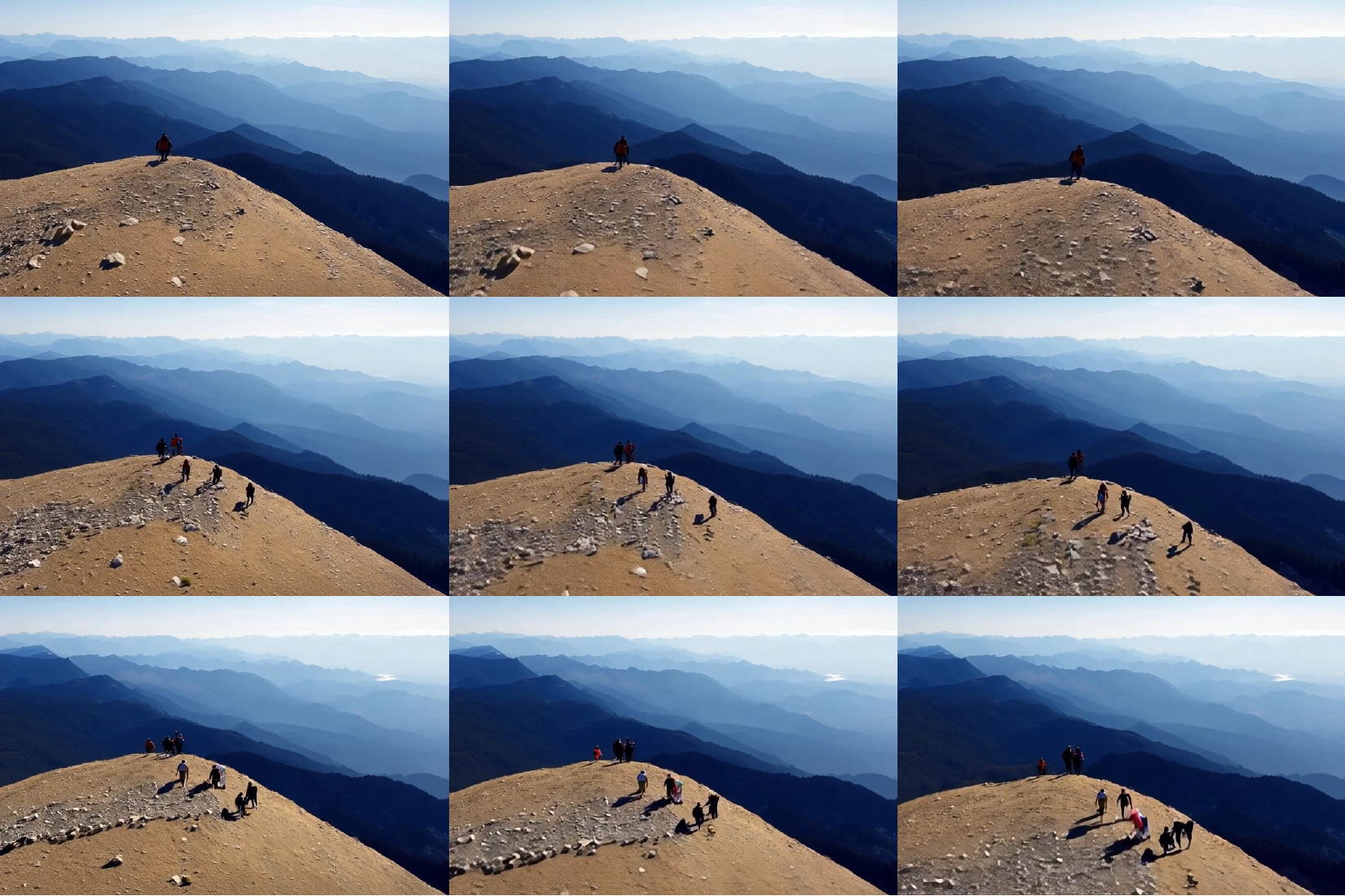}
    \put(35,70){\colorbox{white}{\textcolor{black}{CogVideoX}}}
    \end{overpic}
\end{subfigure}\hfill
\begin{subfigure}{0.535\linewidth}
    \centering
    \begin{overpic}[width=\linewidth]{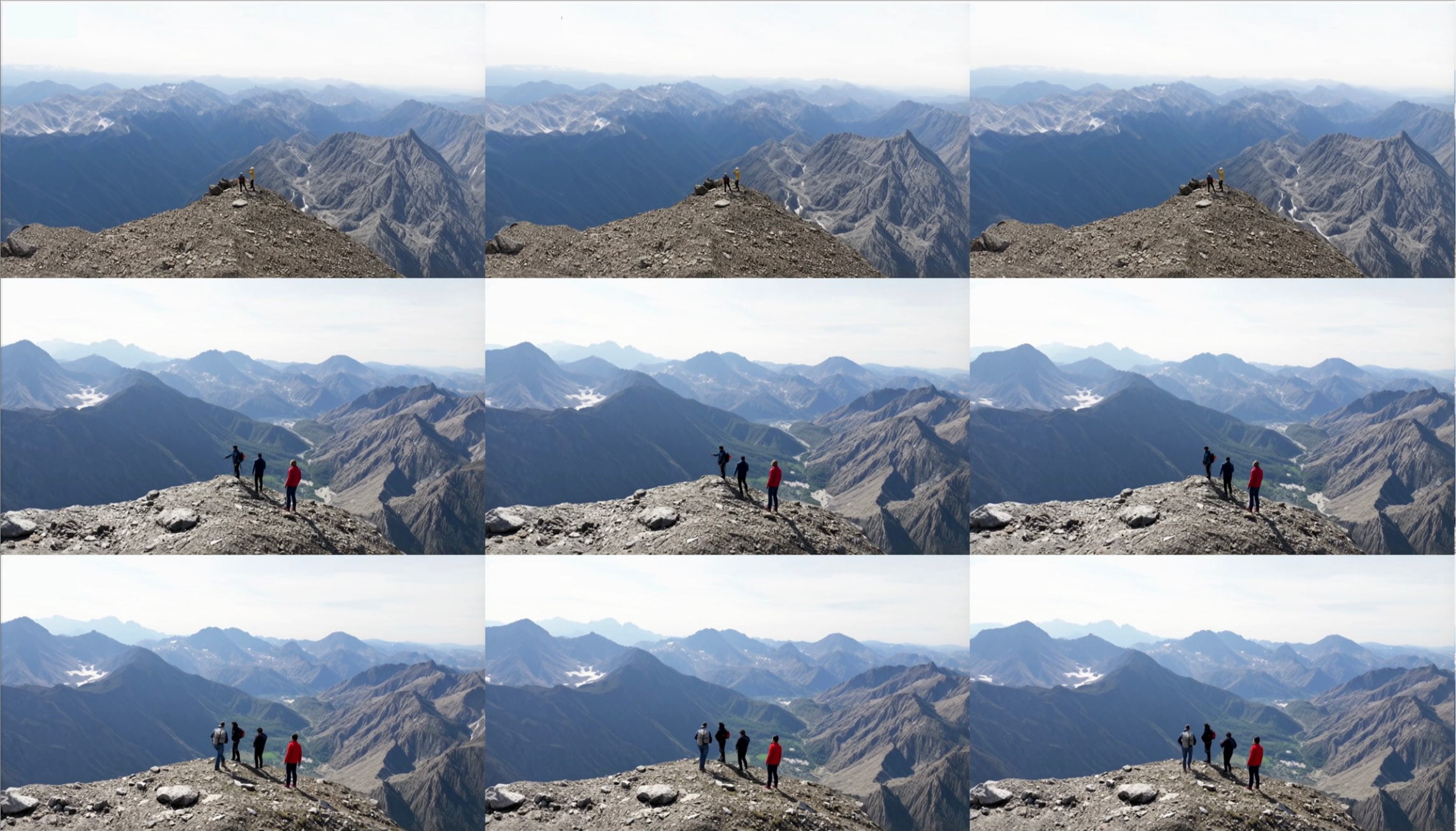}
    \put(38, 59){\colorbox{white}{\textcolor{black}{LTX-Video}}}
    \end{overpic}
\end{subfigure}
\caption{Video sliders for the concept ``mountain hikers'' with different backbones.}
\label{fig:video_different_backbones}
\end{figure}

\paragraph{Qualitative comparison across backbones.}
Modality-agnostic methods like ours are well-suited to the rapidly evolving landscape of foundation models. To qualitatively demonstrate adaptability, we compare the LTX-Video~\citep{hacohen2024ltxvideorealtimevideolatent} vs. CogVideoX (video)~\citep{yang2024cogvideox} and Stable Diffusion v1.4 vs. SD~3 (image). Notably, adopting our method on a new backbone requires only minor adjustments to the sampling procedure. We present a comparison in Fig.~\ref{fig:video_different_backbones} and App.~\ref{sec:qual_comparisons}.

\paragraph{Performance analysis and ablation.} Training CS is expensive in both peak memory and wall clock time (WCT), as seen in Tab.~\ref{tab:complexity}. In comparison, our method requires no training and operates at inference time only. Relative to CS inference, our approach achieves similar memory usage while incurring about $40\%$ more computation due to additional network evaluations. Finally, to analyze ASTD, we recall that it requires generating a small set of samples to run its two-stage procedure. In our setup, we evaluate each concept at 13 scales and generate 30 samples, i.e., an \emph{inference} cost equivalent to 390 sample generations. Thus, the optional ASTD module further boosts performance but comes at the cost of higher runtime and memory: running ASTD is slower than CS training and requires slightly more memory, yet it can be applied \emph{on top of} either method to yield superior results. Specifically, we validate that ASTD consistently improves overall scores across modalities and methods, demonstrating robustness to diverse setups. Tab.~\ref{tab:astd_abl} summarizes the results: the left panel reports CS and our method \emph{without} ASTD, while the right panel shows that all overall scores (OS) increase when ASTD is enabled by nearly $2\times$ in images (I) and audio (A).

\begin{table}[t]
    \centering
    \begin{minipage}{0.48\linewidth}
        \centering
        \caption{Runtime and memory comparisons of CS, our approach, and the ASTD add-on.}
        \label{tab:complexity}
        \begin{tabular}{@{}lccc|c@{}}
        \toprule
         & CS-Train & CS-Infer & Ours & ASTD \\
        \midrule
        WCT (s/m)   & 22 (m) & 6.5 (s) & 9.5 (s) & 31 (m)  \\
        Memory (GB) & 6.7 & 4.4 & 5.0 & 6.8 \\
        \bottomrule
        \end{tabular}
    \end{minipage}\hfill
    \begin{minipage}{0.48\linewidth}
        \centering
        \caption{ASTD contribution ablation.}
        \label{tab:astd_abl}
        \begin{tabular}{@{}lcc|cc@{}}
        \toprule
         & Ours {\footnotesize(w/o)} & CS {\footnotesize(w/o)} & Ours {\footnotesize(w/)} & CS {\footnotesize(w/)} \\
        \midrule
        OS (I) $\uparrow$ & 3.04 & 2.26 & \textbf{3.56} & \textbf{3.11} \\
        OS (V) $\uparrow$ & 2.48 & 2.28 & \textbf{2.70} & \textbf{2.29} \\
        OS (A) $\uparrow$ & 1.83 & 1.82 & \textbf{2.58} & \textbf{2.20} \\
        \bottomrule
    \end{tabular}
\end{minipage}
\end{table}

\begin{figure}[htpb!]
  \centering
  \includegraphics[width=0.9\linewidth]{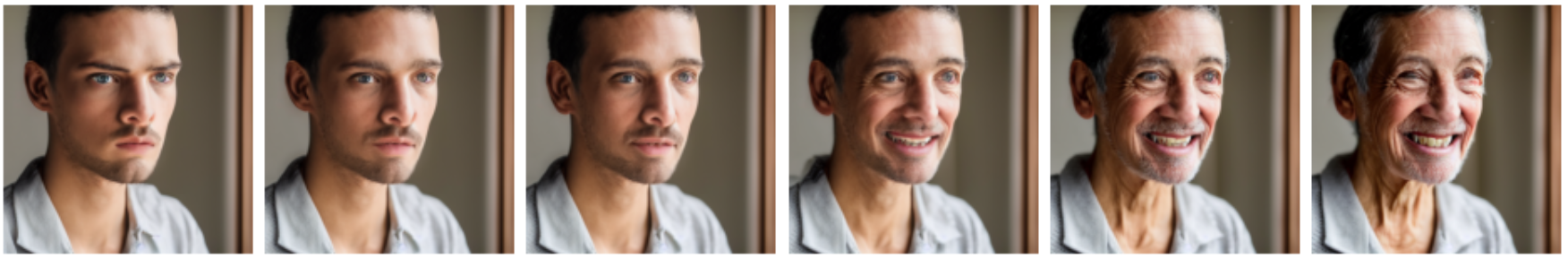}
  \caption{Slider composition: the concepts ``age'' and ``smile'' combined together.}
  \label{fig:slider_composition}
\end{figure}

\paragraph{Slider composition.}
Our method naturally supports compositional edits, enabling multiple sliders to be applied jointly. As shown in Fig.~\ref{fig:slider_composition}, we simultaneously adjust the \emph{age} and \emph{smile} concepts, demonstrating controlled manipulation across multiple attributes.

\section{Discussion}
In this work, we demonstrate that fine-grained concept control in diffusion models can be achieved \emph{without} per-concept training or architecture-specific modifications. By directly estimating the Concept Slider update during inference, our method enables plug-and-play, modality-agnostic sliders across images, video, and audio. This design trades a modest increase in inference cost for substantial gains in generality, ease of deployment, and performance, surpassing strong baselines in controllable generation. Nevertheless, several limitations remain. Long sequences (e.g., video or audio) introduce higher computational demands, alignment models such as CLIP, ViCLIP, and CLAP may propagate biases, and our two-stage saturation detection requires additional samples. These challenges highlight opportunities for improvement through inference-time acceleration (e.g., few-step solvers, distillation), better handling of alignment biases, and more efficient saturation modeling.

Looking forward, promising directions include human-in-the-loop traversal shaping, principled multi-concept disentanglement and composition, and extensions to emerging modalities such as 3D, robotics, and cross-lingual generation. More broadly, lowering technical barriers to controllable generation has the potential to democratize creative and scientific workflows, but it also raises concerns regarding misuse and fairness. Standardized, transparent benchmarks and metrics, combined with safeguards for sensitive attributes, are essential to ensure that training-free, architecture-independent concept sliders are deployed responsibly, interpretably, and ethically.

\clearpage
\bibliography{refs}
\bibliographystyle{iclr2026_conference}

\clearpage
\appendix

\section{Experimental Details}
\label{sec:exp_details}

\subsection{Algorithm Pseudo Code}
\label{sec:psuedo_code}

\begin{algorithm}[ht]
\caption{Our Inference-Time Method with Subroutines}
\label{alg:guided_diffusion}
\begin{algorithmic}[1]
\Require Initial latents $x_0$, prompt pair $(c_\text{neutral}, c_+, c_-)$, timesteps $T$, 
scales $S$, intervention timestep $k$, guidance scales $g_1$

\State Initialize noise scheduler with $T$ steps
\Comment{Step 1: Diffuse using neutral prompt until $k$}
\For{$t = 0$ to $k-1$}
    \State $\epsilon \gets \text{PredictNoise}(x, c_\text{neutral}, g_1, \text{use\_cfg=True})$
    \State $x \gets \text{StepScheduler}(x, \epsilon, t)$
\EndFor

\Comment{Step 2: Guided diffusion for each scale}
\For{each scale $\mu$ in $S$}
    \State $x_\mu \gets x$
    \For{$t = k$ to $T$}
        \State $\epsilon_\text{neutral} \gets \text{PredictNoise}(x_\mu, c_\text{neutral}, g_1, \text{use\_cfg=True})$
        \State $\epsilon_+ \gets \text{PredictNoise}(x_\mu, c_+, \text{use\_cfg=False})$
        \State $\epsilon_- \gets \text{PredictNoise}(x_\mu, c_-, \text{use\_cfg=False})$
        \State $\epsilon_\text{mod} \gets \epsilon_\text{neutral} + \mu \cdot (\epsilon_+ - \epsilon_-)$
        \State $x_\mu \gets \text{StepScheduler}(x_\mu, \epsilon_\text{mod}, t)$
    \EndFor
\EndFor

\Comment{Subroutine: Predict Noise}
\Procedure{PredictNoise}{$x$, $c$, $\gamma$=None, use\_cfg=False}
\State $x_{\text{model}} \gets \begin{cases} \text{duplicate}(x), & \text{if } \text{use\_cfg} = \text{True} \\ x, & \text{otherwise} \end{cases}$
\State $\epsilon_\theta \gets \text{ModelPredict}(x_\text{model}, c, t, \text{scheduler})$
    \If{use\_cfg = True}
        \State Split $\epsilon_\theta$ into $(\epsilon_\text{uncond}, \epsilon_\text{text})$
        \State $\epsilon_\theta \gets \epsilon_\text{uncond} + \gamma \cdot (\epsilon_\text{text} - \epsilon_\text{uncond})$
    \EndIf
    \State \textbf{return} $\epsilon_\theta$
\EndProcedure
\label{alg:psuedo_method}
\end{algorithmic}
\end{algorithm}

\subsection{Benchmarking Data}
This section contain all the prompts used for quantitative results.

\subsubsection{Images}

\begin{lstlisting}[language=yaml]
- base: "A realistic image of a person."
  positive: "A realistic image of a person, smiling widely, very happy."
  negative: "A realistic image of a person, frowning, very sad."

- base: "A realistic image of a person."
  positive: "A realistic image of a person, very old, aged, wrinkly."
  negative: "A realistic image of a person, detailed facial features, clear skin."

- base: "A realistic image of a car."
  positive: "A realistic image of a car, damaged, broken headlights, dented car, with scrapped paintwork."
  negative: "A realistic image of a car, mint condition, brand new, shiny."

- base: "A realistic image of a room."
  positive: "A realistic image of a room, cluttered, disorganized, dirty, jumbled, scattered."
  negative: "A realistic image of a room, super organized, clean, ordered, neat, tidy."

- base: "A realistic image of a person."
  positive: "A realistic image of a person, with big bushy beard."
  negative: "A realistic image of a person, clean shaven."

- base: "A realistic image of a person."
  positive: "A realistic image of a person, wearing glasses."
  negative: "A realistic image of a person, clear face."
  
- base: "A realistic image of a person."
  positive: "A realistic image of a person, very fat, chubby, overweight, obese."
  negative: "A realistic image of a person, very skinny, thin, slim, lean."

- base: "A realistic image of a person."
  positive: "A realistic image of a person, with makeup, cosmetic, concealer, mascara, lipstick."
  negative: "A realistic image of a person, bare face."

- base: "A realistic image of a person."
  positive: "A realistic image of a person, with shocked look, surprised, stunned, amazed."
  negative: "A realistic image of a person, dull, uninterested, bored, incurious."
  
- base: "A realistic image of a table of food."
  positive: "A realistic image of a table of food, cooked and prepped, in dishes."
  negative: "A realistic image of a table of food, raw, natural, not prepped in any way.
\end{lstlisting}

\subsubsection{Video}

\begin{lstlisting}[language=yaml]
- base: "A white sailboat in the ocean."
  positive: "A white sailboat in the stormy wavy ocean."
  negative: "A white sailboat in the calm flat ocean."

- base: "A river flowing through the valley."
  positive: "A river rushing rapidly through the valley, with very big waves."
  negative: "A river gently trickling through the valley, with very calm flat water."

- base: "A butterfly on a leaf."
  positive: "A butterfly flapping its wings very fast."
  negative: "A butterfly standing on a leaf not moving."

- base: "A silhouette of a person walking on the beach."
  positive: "A silhouette of a person jogging very fast on the beach."
  negative: "A silhouette of a person strolling slowly down the beach."
  
- base: "A mountaintop from afar with a few people hiking down it."
  positive: "A mountaintop from afar with a lot of people hiking down it, very busy."
  negative: "A mountaintop from afar with a no people hiking down it, empty."

- base: "A car cruising down the street."
  positive: "A shiny sports car cruising down the street."
  negative: "An old fashioned faded car cruising down the street."

- base: "A large mountain in the distance."
  positive: "An all white snow-capped large mountain in the distance."
  negative: "A brown dry and rocky large mountain in the distance."

- base: "A tall tree in a windy park."
  positive: "A tall green leafy tree in a windy park, very green and with lots of leaves."
  negative: "A tall dry brown bare tree with in a windy park, no leaves on branches."

- base: "A cat sitting on a windowsill licking its paws."
  positive: "A very fat fluffy, well-groomed cat sitting on a windowsill licking its paws."
  negative: "A very skinny, thin, scruffy cat sitting on a windowsill licking its paws."
  
- base: "A lighthouse in the sea, surrounded by water."
  positive:  "A lighthouse in the sea, surrounded by dark splashy water with lots of waves."
  negative:  "A lighthouse in the sea, surrounded by bright calm, flat water."

\end{lstlisting}

\subsubsection{Audio}
\begin{lstlisting}[language=yaml]
- base: "A cat meowing"
  positive: "A cat meowing loudly"
  negative: "A cat meowing softly"
    
- base: "Ocean wave sounds"
  positive: "Stormy ocean wave sounds"
  negative: "Tranquil ocean wave sounds"

- base: "Car engine sound"
  positive: "Revving car engine"
  negative: "Idling car engine"
    
- base: "Choir singing a classical piece"
  positive: "Choir singing a classical piece in a high pitch"
  negative: "Choir singing a classical piece in a low pitch"
    
- base: "Door slamming shut"
  positive: "Wooden door slamming shut"
  negative: "Metal door slamming shut"

- base: "Crowd clapping"
  positive: "Crowd clapping with echo"
  negative: "Crowd clapping with no echo"

- base: "Solo guitar"
  positive: "Solo electric guitar"
  negative: "Solo acoustic guitar"

- base: "Dog barking"
  positive: "Big dog barking"
  negative: "Small dog barking"

- base: "Rain falling"
  positive: "Heavy rain falling"
  negative: "Light rain falling"

- base: "Crowd of people talking"
  positive: "Crowd of people shouting"
  negative: "Crowd of people wispering"

\end{lstlisting}

\subsection{Baselines Implementation Details}
\subsubsection{Text Embedding Variant}
\label{sec:text_emb}

As an alternative to our inference-time method, we explore a variant that directly manipulates the different text embeddings, which we refer to as Text-Embeds (TE). Given text encoder outputs for the neutral, positive, and negative prompts, we construct modified embeddings by linear interpolation in embedding space. At step $k$ of the slider, the manipulated embedding is defined as
$
e_{mod} = e_{\text{neutral}} + \mu \,(e_{\text{positive}} - e_{\text{negative}}),
$
where $\mu$ represents the scale in the slider. Until step $k$, the diffusion process is conditioned on the neutral embedding $e_{\text{neutral}}$, and from step $k$ onward the manipulated embedding $e_{mod}$ is used for the full denoising process. In contrast, our inference-time method requires three forward passes, one for each embedding (neutral, positive, and negative), after which the manipulation is applied directly to the noise predictions. While TE is computationally simpler, it is also less expressive, as the conditioning signal remains static once $e_{mod}$ is fixed. In practice, we find that TE can yield smooth semantic progressions but often lacks the fine-grained control and robustness of our main approach.
\begin{equation}
\epsilon(x_t,t) \;=\;
\begin{cases}
\epsilon_{\theta}(x_t, e_{neutral}, t), & \text{if } t \le k,\\[4pt]
\epsilon_\theta(x_t,e_{mod}, t), & \text{if } t > k.
\end{cases}
\end{equation}

\subsubsection{Prompt to prompt}

To implement Prompt-to-Prompt (P2P) editing, we follow the original framework for controlling cross-attention maps during diffusion. Given a neutral prompt and an edited prompt (e.g., “a person” vs. “a smiling person”), we first generate a base image with the neutral prompt and record its latent representation and attention maps using an attention controller. During editing, the diffusion process is initialized from the neutral latent, and the cross-attention activations of the edited prompt are gradually interpolated with those of the neutral prompt over the denoising steps. To constrain edits to specific semantic regions, we employ LocalBlend, which constructs spatial masks by pooling attention maps of selected target words and restricting modifications to those areas. Additionally, AttentionReweight amplifies the contribution of selected target tokens (e.g., “smiling”) by scaling their attention weights with an equalizer, enabling smooth variation of the edit along a slider. In practice, we apply a refinement controller (AttentionRefine) to align tokens between the prompts, followed by AttentionReweight with different scales to generate images at varying interpolation strengths. By reusing the latent from the neutral prompt, the identity and overall structure of the original image are preserved, while the desired semantic attributes are gradually introduced. All modifications are taken directly from the original P2P codebase. In our experiments, we observed that for some concepts, the images remain effectively unchanged along the slider despite the attention manipulations.

\subsubsection{Prompt Sliders}

The PromptSliders baseline is based on a training-based textual inversion approach, where a dedicated embedding is learned for each concept. During training, only this new concept embedding is updated, while the rest of the model, including the VAE, U-Net, and most of the text encoder, remains fixed. Once trained, the learned embedding can be used at inference time by inserting it into the text encoder, allowing the model to generate images conditioned on the concept. By adjusting the contribution of the learned embedding through a slider, the approach enables continuous and controllable manipulation of the concept in the generated images, providing a flexible way to explore variations without modifying the underlying diffusion model.

\subsubsection{Concept Sliders}

The Concept Sliders baseline implements continuous control over a learned concept using LoRA adapters on top of a frozen base model. In this approach, a low-rank adapter (LoRA) is trained on a specific concept, modifying only a subset of the transformer's attention layers (to\_q, to\_k, to\_v, to\_out.0) while keeping the rest of the model fixed. During training, for each concept, batches of latent images are generated and denoised using the base model conditioned on three types of prompt embeddings: neutral, positive, and negative. The training objective is a mean-squared error between the predicted noise under a target prompt with the LoRA applied and a combination of predicted noises under neutral, positive, and negative prompts without the LoRA. Formally, if $\epsilon_\theta(\mathbf{x}_t, t, c)$ denotes the noise predicted by the transformer at timestep $t$ for latents $\mathbf{x}_t$ conditioned on text embeddings $c$, the target latent is computed as
$\epsilon_{\text{target}} = \epsilon_\theta(\mathbf{x}_t, t, c_\text{neutral}) + \big( \epsilon_\theta(\mathbf{x}_t, t, c_+) - \epsilon_\theta(\mathbf{x}_t, t, c_-) \big)$,
and the loss minimizes $\text{MSE}(\epsilon_{\text{pred}}, \epsilon_{\text{target}})$ where $\epsilon_{\text{pred}}$ is the LoRA-modified prediction. Once trained, these LoRA weights can be loaded at inference to manipulate the concept continuously. During inference, for a given latent initialization $\mathbf{x}_0$ and scale $\mu$, the noise prediction is computed as
\begin{equation}
    \epsilon_t =
    \begin{cases}
        \epsilon_\theta(x_t, c) , & t > t_0 \\[2mm]
        \epsilon_{\theta + \mu \omega}(x_t, c) , & t \le t_0
    \end{cases} \ ,
\end{equation}
where $\epsilon_\theta$ is the denoising network of the base model, $\omega$ are the concept-specific LoRA weights, and $\mu$ is a scaling factor controlling the intensity of the concept and $t_0$ is the diffusion timestep at which the LoRA-modified weights begin to be applied, while all steps before $t_0$ use the original model weights.

\subsection{Metrics Implementation Details}
\label{sec:app_metrics}

Each metric uses either the perceptual similarity score $d(x_i, x_j)$, which measures the closeness between generated outputs, or the semantic alignment score $a(x, c)$, which quantifies how well an output matches the input prompt. The metrics that build on these scores are described in section \ref{sec:sliders_metrics}, while the exact definition of $d$ and $a$ depends on the modality and is described in the following sections.

\subsubsection{Image}
The perceptual similarity, $d(x_\mu, x_0)$, is calculated using the Learned Perceptual Image Patch Similarity (LPIPS) metric~\cite{zhang2018unreasonable}, which compares the reference image $x_0$ (corresponding to the neutral scale) with a manipulated image $x_\mu$ at scale $\mu$. LPIPS captures perceptual differences between images, reflecting visual fidelity and preserving global structure. 

For semantic alignment $a(x, c)$, we compute the CLIP-based score~\citep{radford2021learning}, which measures the cosine similarity between the image embedding of $x$ and the text embedding of the corresponding prompt $c$.

\subsubsection{Video}
In video, both static and dynamic aspects need to be measured, which we capture by defining separate scores for each. The \emph{static} perceptual similarity $d(S)(x_\mu, x_0)$ is measured using LPIPS~\citep{zhang2018unreasonable} with an AlexNet backbone~\citep{voleti2022mcvd}, which primarily captures more static differences between frames, reflecting changes in identity or appearance. \emph{Dynamic} differences between videos $d(D)(x_\mu, x_0)$, like movement or speed, is measured using \emph{Motion Alignment}~\citep{xu2022gmflow}, which yields a score sensitive to dynamic changes in motion patterns.

Semantic alignment with the target concept is assessed using text-video CLIP-based scores. \emph{CLIP-Text Alignment}~\citep{radford2021learning} computes the similarity between individual frame embeddings and the textual prompt embedding, capturing more static semantic consistency, denoted by $a(S)(x, c)$, whereas \emph{ViCLIP-Text Alignment}~\citep{wang2023internvid} computes a spatio-temporal embedding of the video and compares it to the text prompt embedding, capturing dynamic semantic alignment across frames, denoted by $a(D)(x, c)$. 

In section \ref{sec:sliders_metrics}, the metrics refer to the scores as $d$ and $a$, for videos, these should be understood as $d(S), a(S)$ for the static versions and $d(D), a(D)$ for the dynamic versions of the metrics.

\subsubsection{Audio}

For generated audio, perceptual similarity, \(d(x_\mu, x_0)\), is computed with the LPAPS metric following \citep{manor2024zeroshot}, using the LAION-AI CLAP library \citep{laionclap2023} with the \texttt{music\_speech\_epoch\_15\_esc\_89.25.pt} weights \citep{SpecVQGAN_Iashin_2021, paissan2023audio}. We compare the reference audio \(x_0\) (neutral scale) to the manipulated audio \(x_\mu\) at manipulation scale \(\mu\).

Semantic alignment is measured with the CLAP score \(a(x, c)\), defined as the cosine similarity between the audio embedding of \(x\) and the text embedding of the corresponding prompt \(c\) \citep{wu2023latent, htsatke2022}. Following \citep{manor2024zeroshot}, we compute CLAP using the LAION-AI CLAP library \citep{laionclap2023} with the \texttt{music\_speech\_audioset\_epoch\_15\_esc\_89.98.pt} weights.

\subsection{Delta Clip} 
\label{sec:delta_clip}

By using the corresponding alignment score $a(x, c)$ for each modality, let $a(x_\mu, c_+)$ denote the CLIP score of the image at scale $\mu$ compared the positive prompt $c_+$.
The $\Delta CLIP$ score for each scale $\mu$ is calculated by
$\Delta \text{CLIP}(\mu) = \big| a(x_\mu, c_+) - a(x_0, c_+) \big|$, and the overall $\Delta CLIP$ for the slider is the average across all non-neutral scales $S$:
$\Delta \text{CLIP}_\text{avg} = \frac{1}{|\mathcal{S} \setminus \{0\}|} \sum_{\mu \in \mathcal{S} \setminus \{0\}} \Delta \text{CLIP}(\mu),  \mu \in S$.

\section{Diffusion Models Implementation Details}

\subsection{Image}
We use Stable Diffusion 1-4~\cite{rombach2022high} for images with resolution $512 \times 512$. The diffusion process consists of $50$ total timesteps, and for the methods where it is relevant a splitting point at timestep $k=15$, and for methods with LoRA weights the starting point to combine the weights is $start noise=800$. The guidance scale used in the diffusion for the neutral prompt is $g_1=7.5$ for classifier-free guidance. Stable Diffusion 1-5 was also used for some qualitative results.

\subsection{Video}
For quantitative results we use CogVideoX-2b~\cite{yang2024cogvideox} with a resolution of $480 \times 720$, with 24 frames. The diffusion process consists of $50$ total timesteps, and for the methods where it is relevant a splitting point at timestep $k=12$, and for methods with LoRA weights the starting point to combine the weights is $start-noise=800$. The guidance scale used in the diffusion for the neutral prompt is $g_1=7.5$ for classifier free guidance.
For qualitive results and comparisons we also used LTX-Video~\cite{hacohen2024ltxvideorealtimevideolatent} as a backbone, with resolution of $1344 \times 768$ with a total of 97 frames. The diffusion consists of 30 steps overall with splitting points tested at $k=10$.

\subsection{Audio}

We use Stable Audio Open 1.0~\cite{EvansPCZTP25} with an implementation based on the \texttt{stable-audio-tools} library and a Heun sampler~\citep{karras2022elucidating}. The sampler hyperparameters are \(\sigma_{\min}=0.3\), \(\sigma_{\max}=500\), \(\text{steps}=36\), \(\rho=3\), \(s_{\text{churn}}=0\), and a guidance scales of \(g_1 = 7\). We generate 10\,s of audio at a 44{,}100\,Hz sampling rate.
For all methods where it is relevant, we branch at \(k=4\).
In our setting, we use \(\epsilon_{\theta}^{\text{mod}}\) in both the first- and second-order steps of the sampler.

\section{Additional Results}

\subsection{Qualitative Experiments}
\label{sec:qual_comparisons}

\begin{figure}[ht]
    \centering
    \begin{subfigure}{1.0\linewidth}
        \centering
        \includegraphics[width=\linewidth]{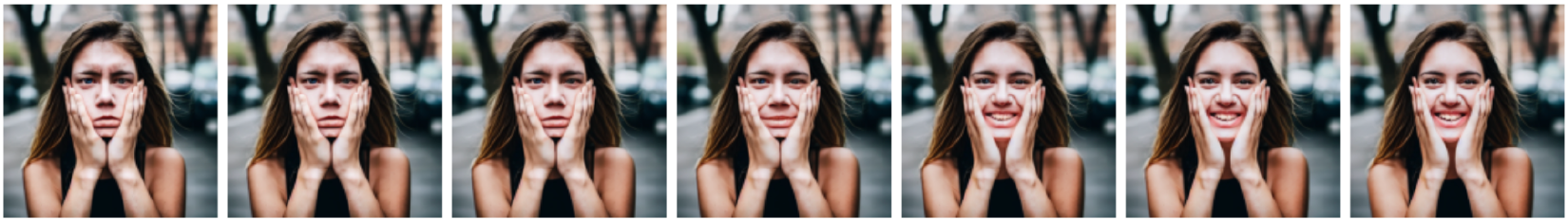}
        \vspace{-5mm}
        \caption{Our inference-time method.}
    \end{subfigure}
    \begin{subfigure}{1.0\linewidth}
        \centering
        \includegraphics[width=\linewidth]{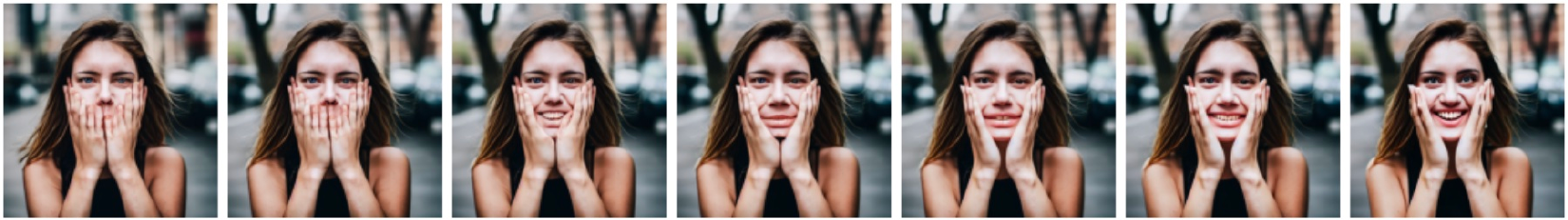}
        \vspace{-5mm}
        \caption{CS method.} 
    \end{subfigure}
    \begin{subfigure}{1.0\linewidth}
        \centering
        \includegraphics[width=\linewidth]{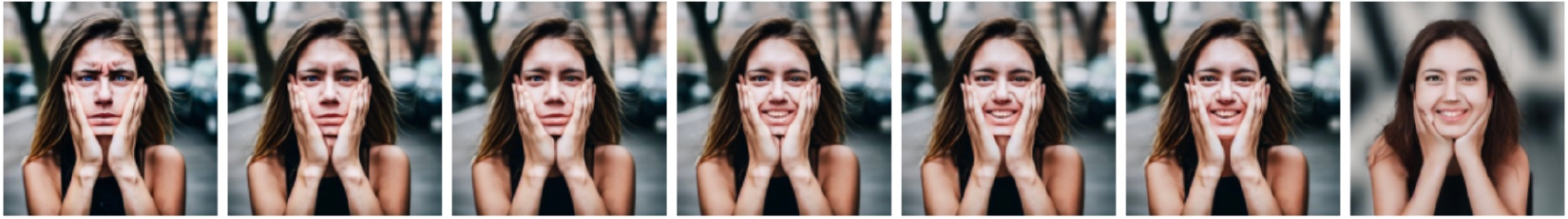}
        \vspace{-5mm}
        \caption{Text Embeddings (TE) variant.}
    \end{subfigure}
    \caption{Image slider of concept ``smiling'' with different methods.}
    \label{fig:comparison_image_methods_smiling}
\end{figure}

\begin{figure}[ht]
    \vspace{-2mm}
    \centering
    \begin{subfigure}{1.0\linewidth}
        \centering
        \includegraphics[width=\linewidth]{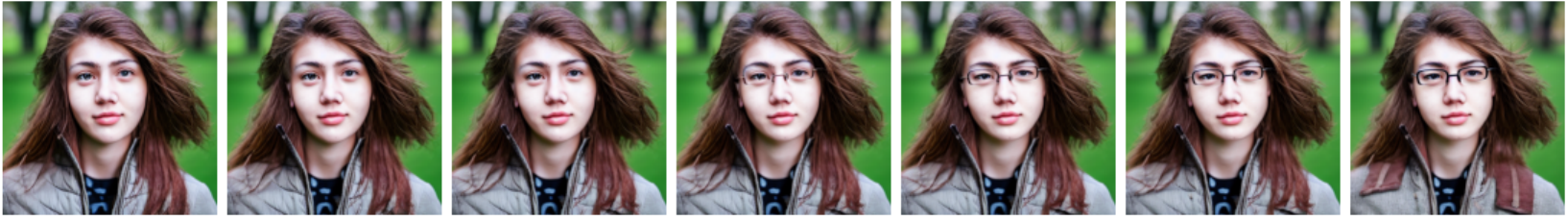}
        \vspace{-5mm}
        \caption{Our inference-time method.}
    \end{subfigure}
    \begin{subfigure}{1.0\linewidth}
        \centering
        \includegraphics[width=\linewidth]{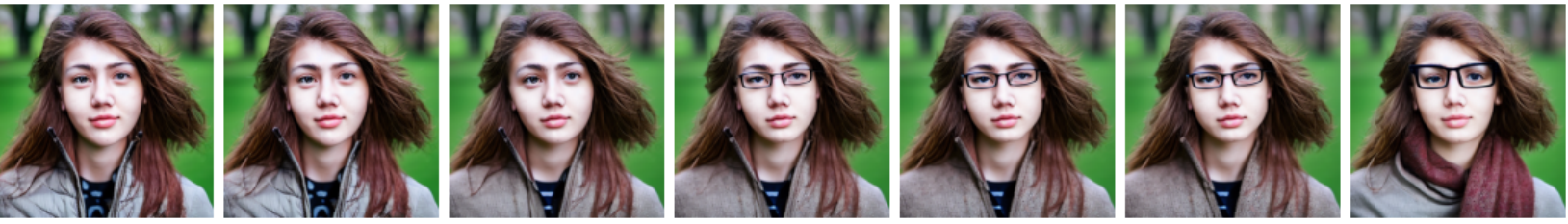}
        \vspace{-5mm}
        \caption{CS method.} 
    \end{subfigure}
    \begin{subfigure}{1.0\linewidth}
        \centering
        \includegraphics[width=\linewidth]{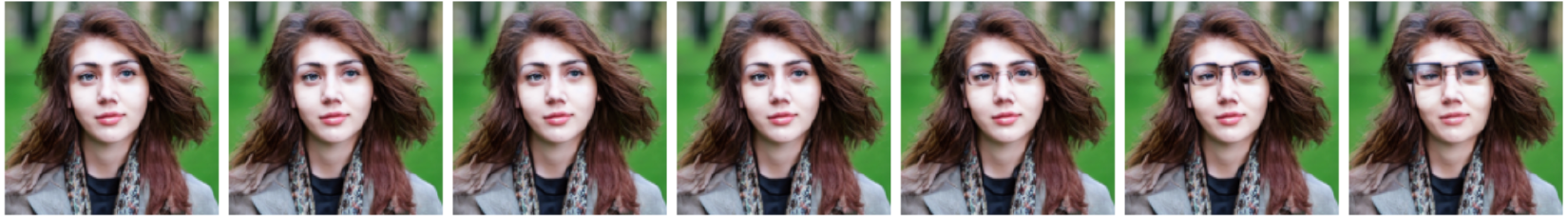}
        \vspace{-5mm}
        \caption{Text Embeddings (TE) variant.}
    \end{subfigure}
    \caption{Image slider of concept ``glasses'' with our inference-time method (top), CS method (middle) and Text-Embeddings variant (bottom).}
    \label{fig:comparison_image_methods_glasses}
\end{figure}

\begin{figure}[ht]
    \centering

    \begin{subfigure}{\linewidth}
        \centering
        \includegraphics[width=1.0\linewidth]{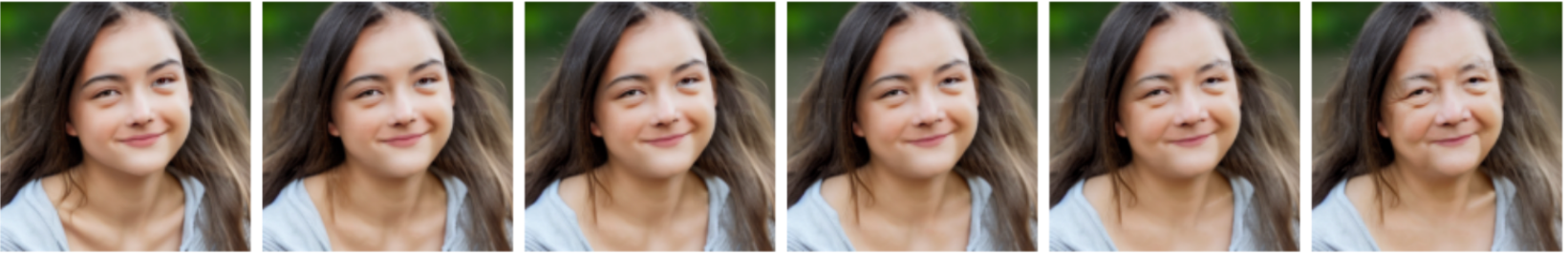}
        \vspace{-5mm}
        \caption{Our method with generic scales.}
    \end{subfigure}
    \begin{subfigure}{\linewidth}
        \centering
        \includegraphics[width=1.0\linewidth]{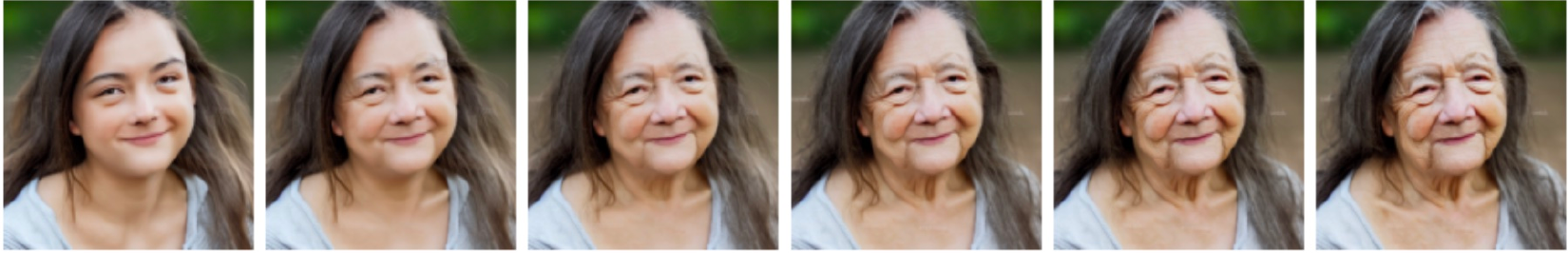}
        \caption{Our method with scales found by ASTD add-on.}
        \vspace{-5mm}
    \end{subfigure}
    
    \vspace{8mm}
    
    \begin{subfigure}{\linewidth}

        \centering
        \includegraphics[width=1.0\linewidth]{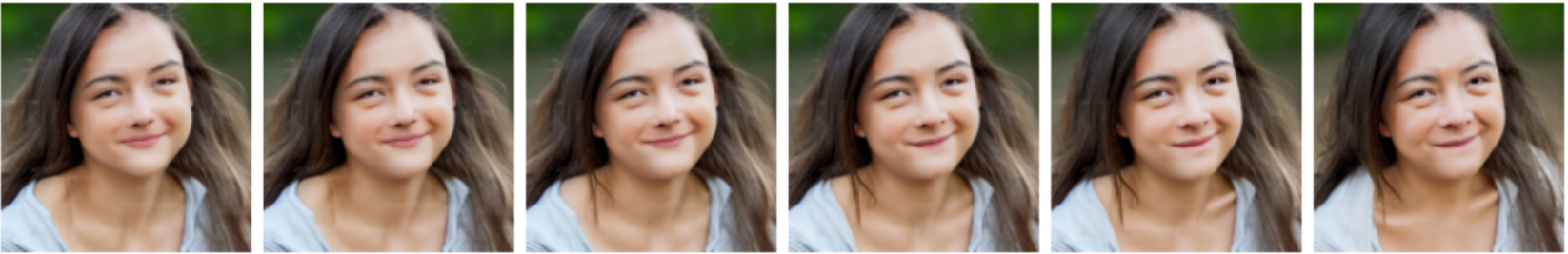}
        \vspace{-5mm}
        \caption{CS with generic scales.}

    \end{subfigure}
    \begin{subfigure}{\linewidth}

        \centering
        \includegraphics[width=1.0\linewidth]{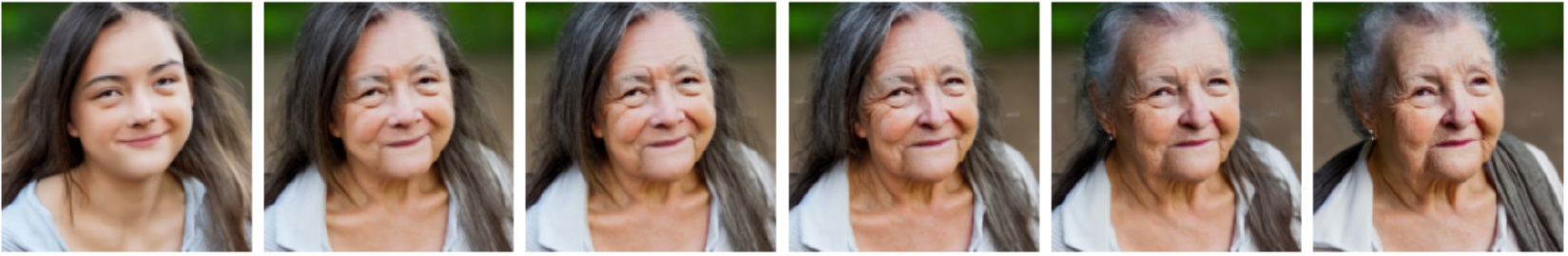}
        \vspace{-5mm}
        \caption{CS with scales found by ASTD add-on.}
    \end{subfigure}
    \caption{Comparison of our method and CS with and without ASTD.}
    \label{fig:comparison_image_astd}
\end{figure}

\paragraph{Images} Examples for comparison between methods can be found in figures \ref{fig:comparison_image_methods_smiling} and \ref{fig:comparison_image_methods_glasses}, and comparisons of methods with and without the ASTD add-on can be found in figure \ref{fig:comparison_image_astd}.

\begin{figure}[htpb!]
    \centering
    \begin{subfigure}{1.0\linewidth}
        \centering
        \includegraphics[width=\linewidth]{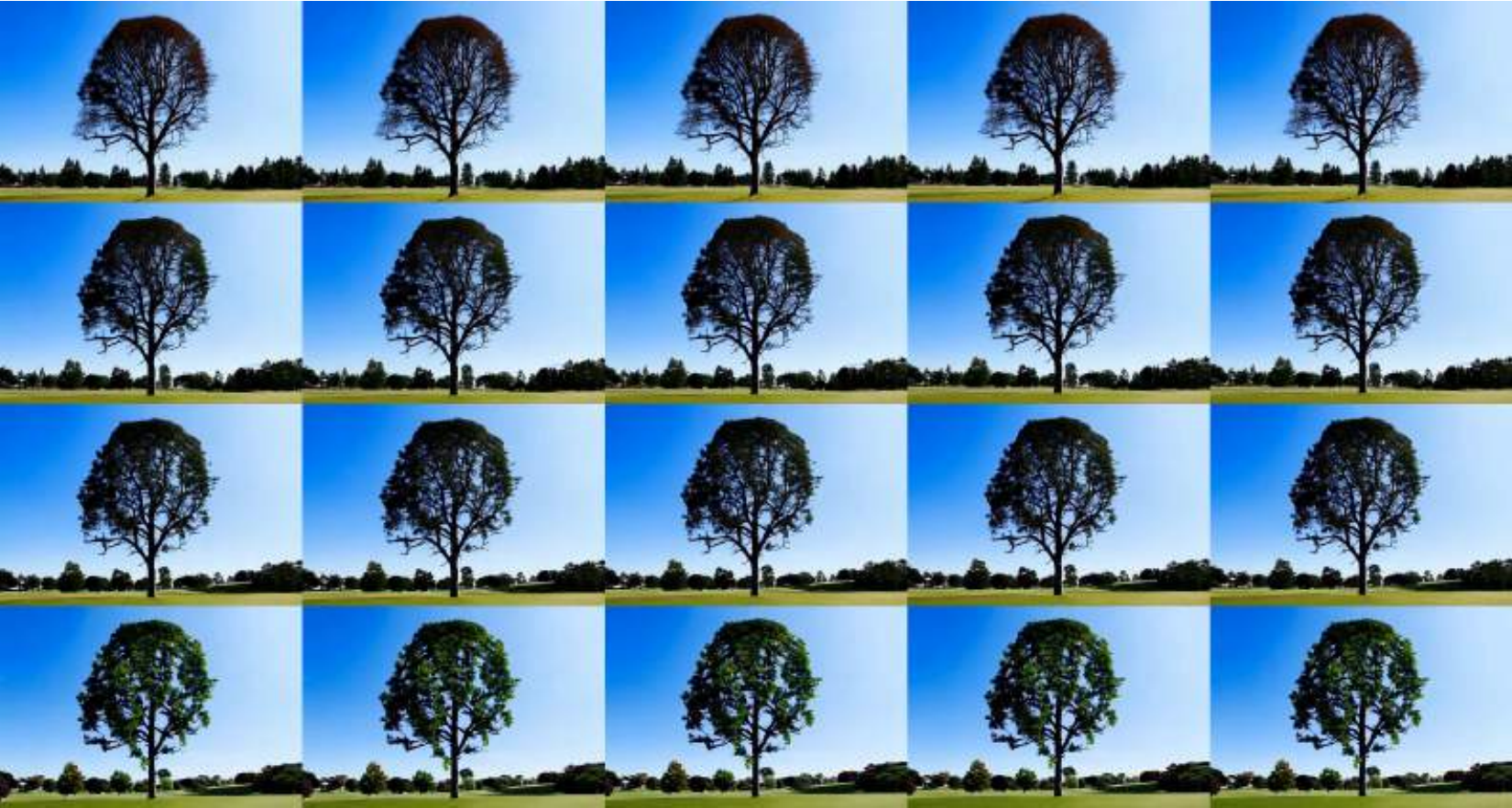}
        \caption{Our method.}
    \end{subfigure}
    \begin{subfigure}{1.0\linewidth}
        \centering
        \includegraphics[width=\linewidth]{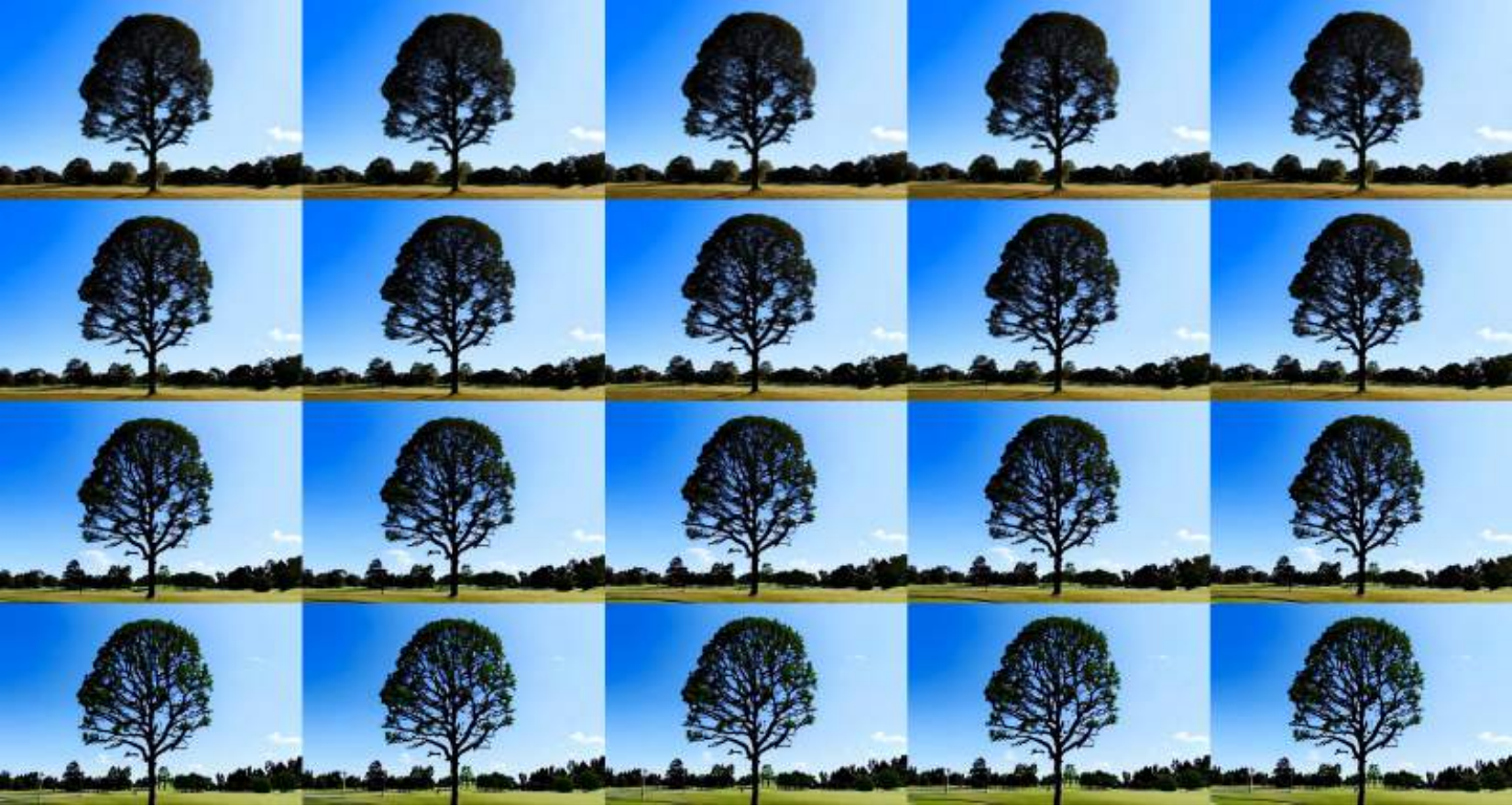}
        \caption{CS method.}
    \end{subfigure}
    \caption{Video Slider with our inference-time method (top) and with CS method (bottom) with CogVideoX backbone over concept ``leafy tree''. The rows depict increasing scales and the columns different frames.}
    \label{fig:comparison_video_tree}
\end{figure}

\begin{figure}[ht]

  \centering
  \includegraphics[width=\textwidth]{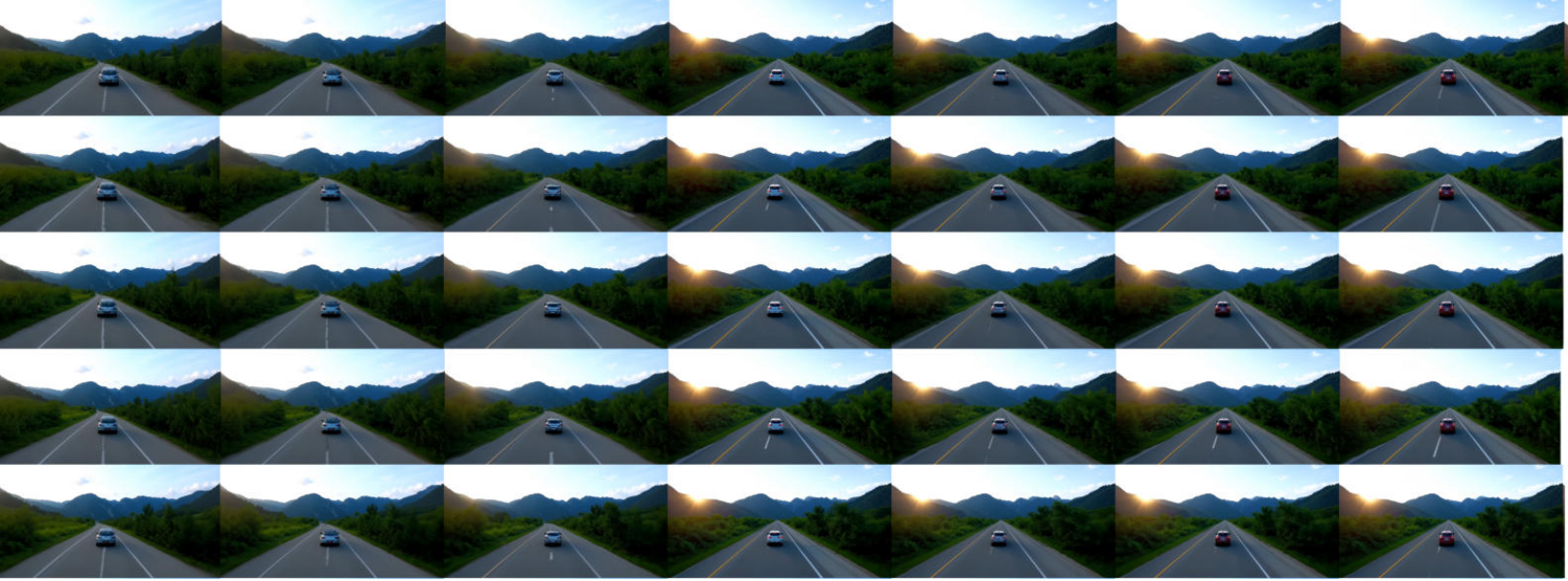}
  \caption{Video slider on concept "car style" computed with LTX-Video as backbone. The columns represent different scales on the slider changing the car style and the rows are the frames of the video.}
  \label{fig:ltx_car_small}
\end{figure}

\begin{figure}[ht]

  \centering
  \includegraphics[width=\textwidth]{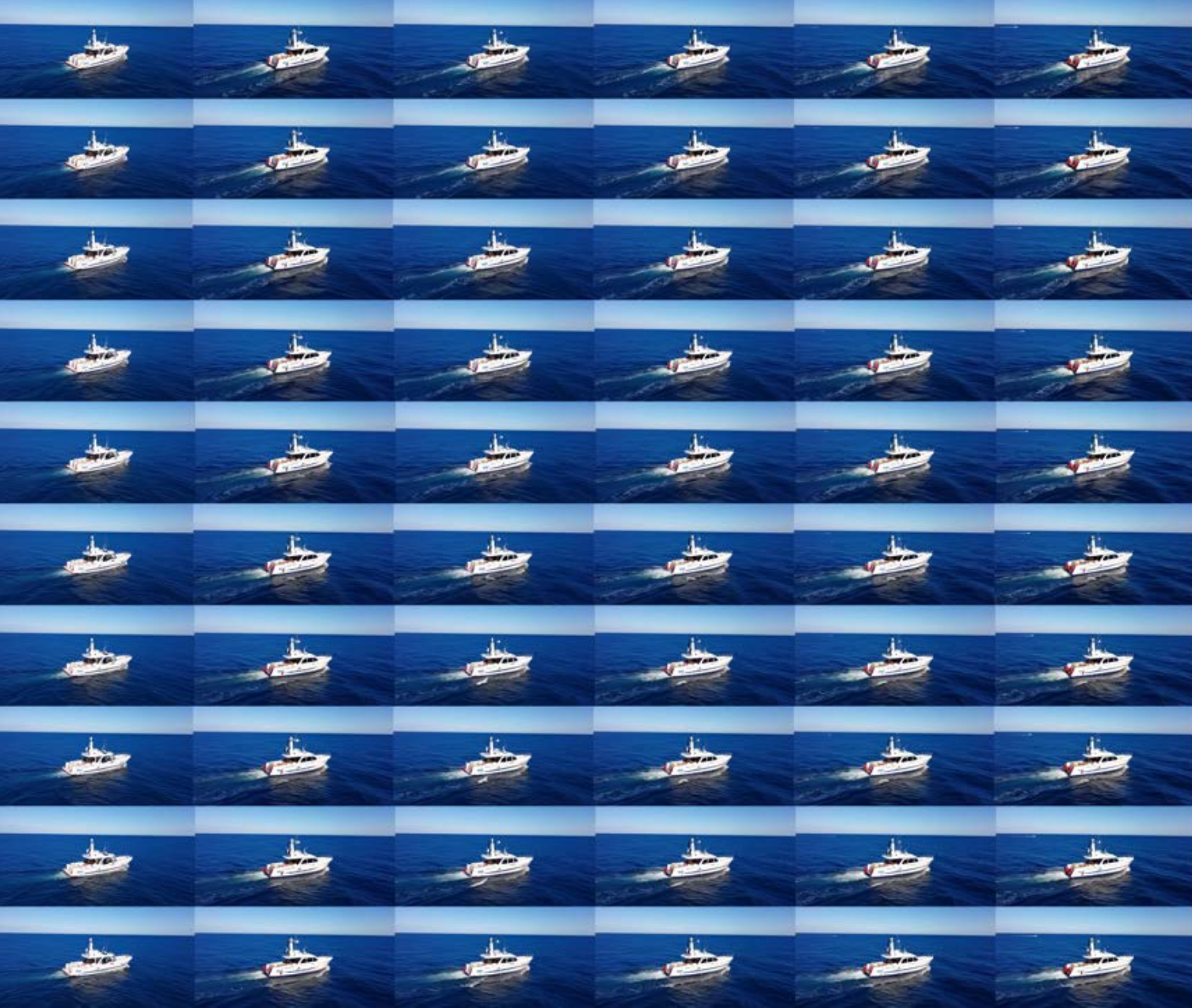}
  \caption{Video slider on concept "sailboat waves" computed with LTX-Video as backbone. The columns represent different scales on the slider changing the waves intensity and the rows are the frames of the video.}
  \label{fig:ltx_sailboat}
\end{figure}

\paragraph{Video} Examples for video sliders with the LTX-Video backbone can be found in Fig. \ref{fig:ltx_car_small}, \ref{fig:ltx_sailboat}. Comparison between our method and CS can be found in Fig.~\ref{fig:comparison_video_tree}.

\subsection{Quantitative Experiments}
\label{sec:quant_comparisons}

\subsubsection{Images}
\paragraph{Concepts:} \emph{Age}, \emph{Chubby}, \emph{Smiling}, \emph{Surprised}, \emph{Lipstick}, \emph{Glasses}, \emph{Beard}, \emph{Car damage}, \emph{Cooked food}, and \emph{Cluttered room}

The results were computed over the 10 concepts, each one generated sliders over 100 different seeds, and with 7 scales. The generic scales (without the ASTD add-on) are [-3, -2, -1, 0, 1, 2, 3], while the scales found with the ASTD add-on vary per concept and per method. See table \ref{tab:all_image_results} for full results for all concepts.

\subsubsection{Video}
\paragraph{Concepts:} \emph{mountain hikers}, \emph{car type}, \emph{mountain type}, \emph{tree leafiness}, \emph{cat size},
\emph{shore waves near a lighthouse}, \emph{ocean waves on a sailboat}, \emph{river current},
\emph{butterfly wing flapping}, and \emph{walking speed on the shore}

For video sliders, the results were computed over 10 concepts, that vary in changing static and dynamic features. each concept was evaluated over 10 different seeds with 7 scales per slider. The generic scales are [-3, -2, -1, 0, 1, 2, 3], while the scales found with ASTD vary for each concept and method. See table \ref{tab:all_video_results} for all results.

\subsubsection{Audio}
\paragraph{Concepts:} \emph{cat meow}, \emph{ocean waves}, \emph{car engine}, \emph{choir (classical)}, \emph{door slam}, \emph{crowd clapping}, \emph{solo guitar}, \emph{dog bark}, \emph{rainfall}, and \emph{people talking}

The results were computed over the 10 concepts, each one generated sliders over 10 different seeds, and with 7 scales. The generic scales (without the ASTD add-on) are [-3, -2, -1, 0, 1, 2, 3], while the scales found with the ASTD add-on vary per concept and per method. See table \ref{tab:all_audio_results} for full results for all concepts.

\subsection{Delta Clip Limitations Cont.}
\label{sec:failure_cases_cont}

\begin{figure}[b!]
\vspace{-1mm}
\centering
\captionsetup[subfigure]{aboveskip=2pt, belowskip=2pt}

\begin{subfigure}{0.90\linewidth}
  \centering
\begin{minipage}{0.08\linewidth}
    \metricbox{black}{3.1}\\[2pt]
    \metricbox{blue!70!black}{2.7}\\[2pt]
    \metricbox{red!70!black}{.27}
  \end{minipage}
  \begin{minipage}{0.90\linewidth}
    \includegraphics[width=\linewidth]{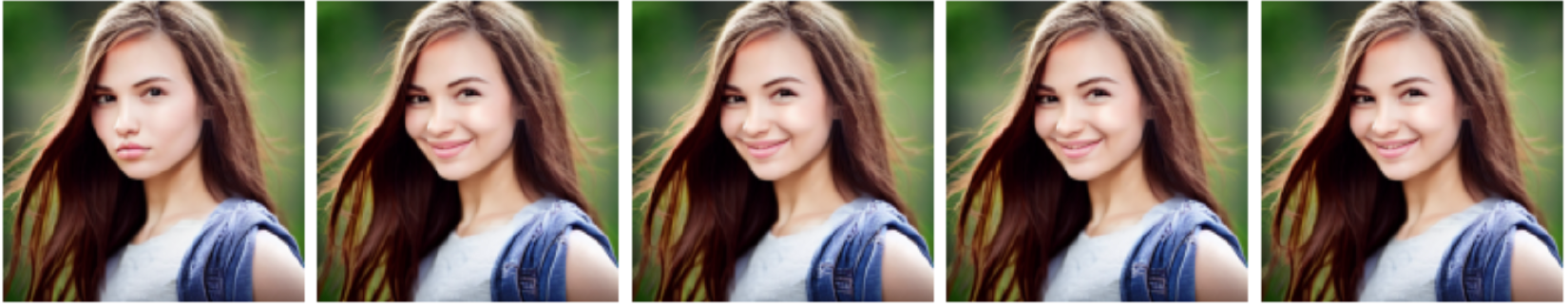}
  \end{minipage}%
\end{subfigure}\hfill
\begin{subfigure}{0.90\linewidth}
  \centering
    \begin{minipage}{0.08\linewidth}
    \metricbox{black}{1.9}\\[2pt]
    \metricbox{blue!70!black}{3.9}\\[2pt]
    \metricbox{red!70!black}{.28}
  \end{minipage}
  \begin{minipage}{0.90\linewidth}
    \includegraphics[width=\linewidth]{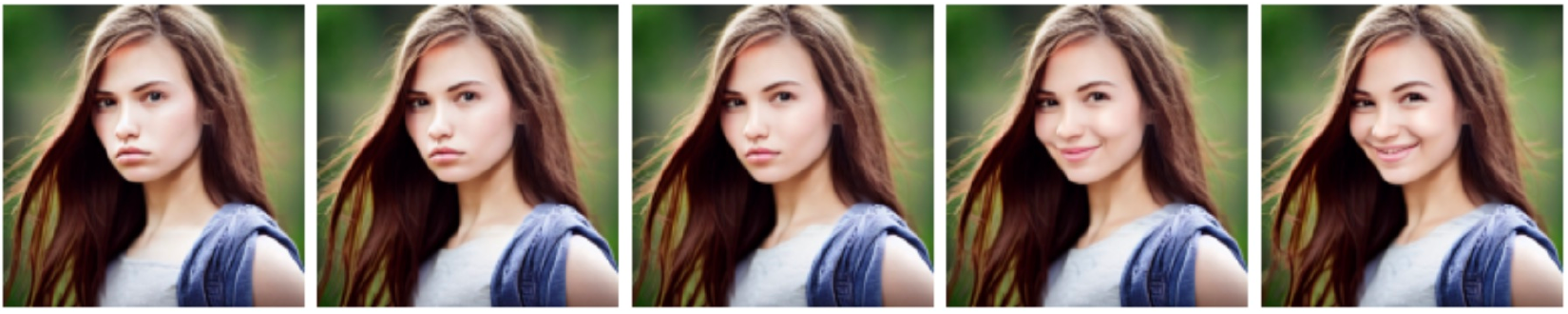}
  \end{minipage}%
\caption{Case where $\Delta$CLIP fails to capture negative range correctly.}
\label{fig:app_dc_failure_range}
\end{subfigure}
\vspace{2mm}
\begin{subfigure}{0.90\linewidth}
  \centering
    \begin{minipage}{0.08\linewidth}
    \metricbox{black}{2.1}\\[2pt]
    \metricbox{blue!70!black}{2.3}\\[2pt]
    \metricbox{red!70!black}{.30}
  \end{minipage}
  \begin{minipage}{0.90\linewidth}
    \includegraphics[width=\linewidth]{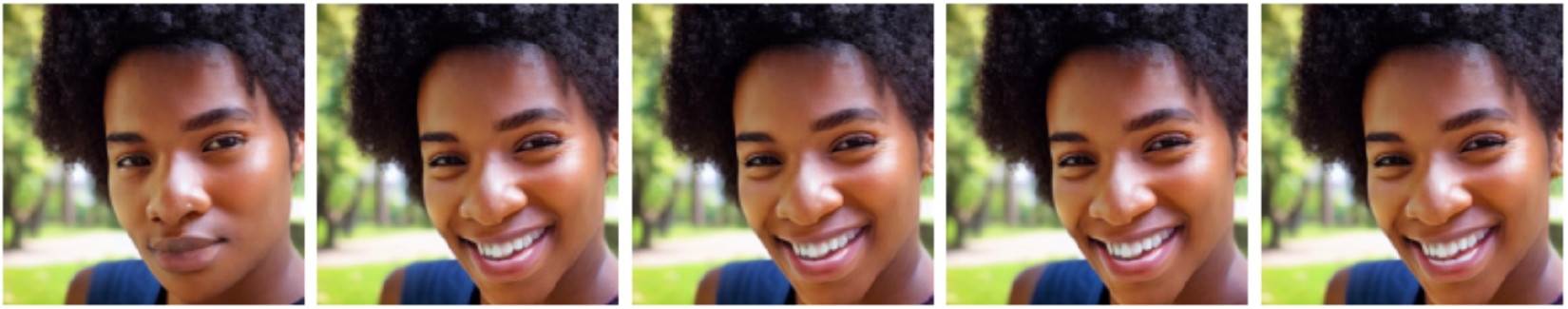}
  \end{minipage}%

\end{subfigure}\hfill
\begin{subfigure}{0.90\linewidth}
  \centering
  \begin{minipage}{0.08\linewidth}
    \metricbox{black}{1.4}\\[2pt]
    \metricbox{blue!70!black}{3.2}\\[2pt]
    \metricbox{red!70!black}{.29}
  \end{minipage}
  \begin{minipage}{0.90\linewidth}
    \includegraphics[width=\linewidth]{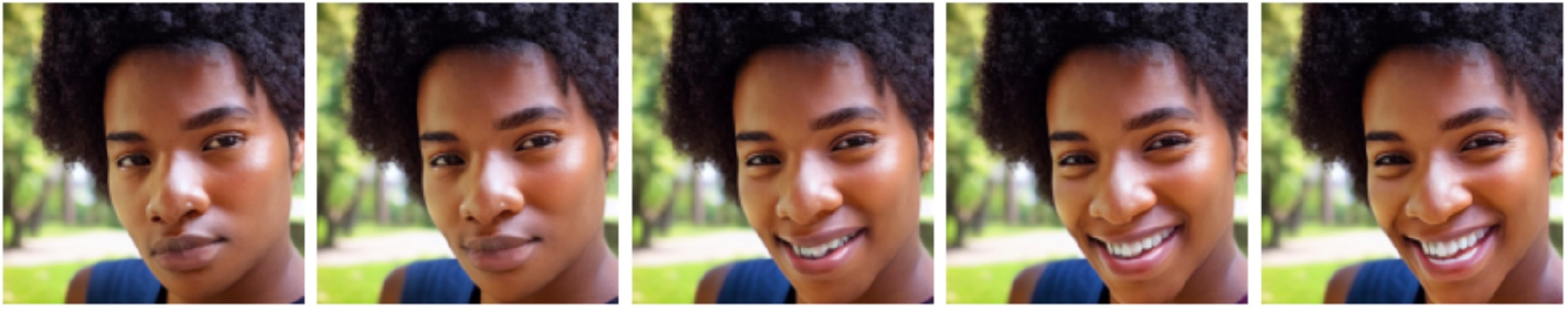}
  \end{minipage}%
\caption{Case where $\Delta$CLIP fails to reflect interval smoothness.}
\label{fig:app_dc_failure_smoothness}
\end{subfigure}
\caption{Examples of limitations for the $\Delta$CLIP metric (black boxes) that CR and CSM metric results (blue and red boxes, respectively) capture.}
\end{figure}

In figure~\ref{fig:metric-failure-cases}, we show several cases where a visually better slider receives a lower $\Delta$CLIP score, revealing its inability to capture the range and smoothness of transitions. One limitation is shown in figure~\ref{fig:app_dc_failure_smoothness}, and occurs when the slider produces an abrupt change at the beginning, after which the images remain constant. Although $\Delta$CLIP assigns a high score due to the large initial jump in CLIP similarity, it doesn't reflect the lack of smooth intermediate progression. Another limitation arises when the change is pronounced only in the negative direction but not in the positive direction, causing the metric to overlook asymmetries and doesn't succeed in capturing both directions, like in figure~\ref{fig:app_dc_failure_range}. 
In contrast, the proposed metrics in Section~\ref{sec:sliders_metrics} capture both range and smoothness, aligning more closely with the visual results.

\subsection{Slider Composition}
\label{sec:slider-composition}
\begin{figure}[htpb!]
  \centering
  \includegraphics[width=\linewidth]{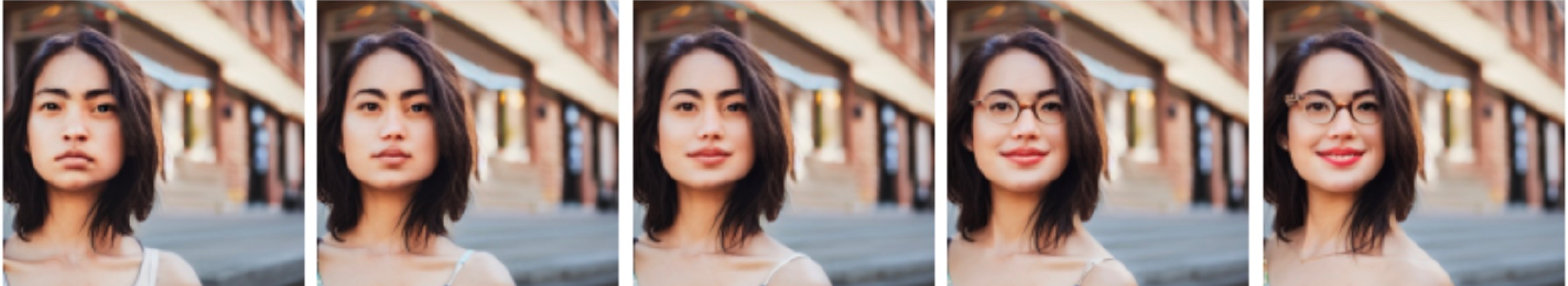}
  \caption{Slider composition: Three concepts ``smile'', ``glasses'' and ``lipstick'' combined together.}
  \label{fig:slider_composition2}
\end{figure}

Our method naturally supports compositional edits. As shown in Fig.~\ref{fig:slider_composition2}, we simultaneously adjust the \emph{smile}, \emph{glasses}, and \emph{lipstick} concepts, demonstrating controlled manipulation across multiple attributes. To create the combination of the concepts, the prompts of every concept are concatenated one after another and are used jointly in the method as a single prompt, forming the basis for the slider composition.

\begin{table*}[!ht]
\centering
\caption{Overall results for images evaluation, for short R=Conceptual Range, S=Smoothness, P=Preservation, the values with a dash are sliders where the images weren't changed along the slider.}
\label{tab:all_image_results}
\resizebox{1\textwidth}{!}{
\begin{tabular}{l|cc|cc|cc|cc|cc}
\toprule
 & \multicolumn{2}{c|}{\textbf{CS}} & \multicolumn{2}{c|}{\textbf{Ours}} & \multicolumn{2}{c|}{\textbf{TE}} & \multicolumn{2}{c|}{\textbf{PromptSliders}} & \multicolumn{2}{c}{\textbf{P2P}} \\
\cmidrule(lr){2-3} \cmidrule(lr){4-5} \cmidrule(lr){6-7} \cmidrule(lr){8-9} \cmidrule(lr){10-11}
 & \textbf{w/o ASTD} & \textbf{w/ ASTD} & \textbf{w/o ASTD} & \textbf{w/ ASTD} & \textbf{w/o ASTD} & \textbf{w/ ASTD} & \textbf{w/o ASTD} & \textbf{w/ ASTD} & \textbf{w/o ASTD} & \textbf{w/ ASTD} \\ \midrule
\multicolumn{11}{c}{\textbf{Concept: Age}} \\ \midrule
R$\uparrow$ & $2.27 \pm 1.24$ & $4.64 \pm 1.08$ & $3.95 \pm 1.39$ & $5.10 \pm 1.37$ & $2.53 \pm 1.30$ & $1.79 \pm 0.62$ & $-0.01 \pm 0.02$ & $0.11 \pm 0.03$ & $0.00 \pm 0.00$ & $0.00 \pm 0.00$ \\
S$\downarrow$ & $0.28 \pm 0.01$ & $0.27 \pm 0.01$ & $0.28 \pm 0.01$ & $0.27 \pm 0.01$ & $0.28 \pm 0.01$ & $0.28 \pm 0.01$ & $0.28 \pm 0.01$ & $0.29 \pm 0.01$ & $0.30 \pm 0.00$ & $0.29 \pm 0.00$ \\
P$\downarrow$ & $0.02 \pm 0.03$ & $0.06 \pm 0.01$ & $0.02 \pm 0.01$ & $0.03 \pm 0.03$ & $0.02 \pm 0.02$ & $0.01 \pm 0.01$ & $0.04 \pm 0.02$ & $0.04 \pm 0.01$ & $0.00 \pm 0.00$ & --- \\ \midrule
\multicolumn{11}{c}{\textbf{Concept: Beard}} \\ \midrule
R$\uparrow$ & $1.69 \pm 1.57$ & $3.59 \pm 0.87$ & $2.12 \pm 1.63$ & $3.63 \pm 1.03$ & $0.97 \pm 1.37$ & $0.28 \pm 1.26$ & $-0.54 \pm 0.05$ & $0.00 \pm 0.00$ & $0.00 \pm 0.00$ & $0.00 \pm 0.00$ \\
S$\downarrow$ & $0.30 \pm 0.01$ & $0.29 \pm 0.01$ & $0.29 \pm 0.01$ & $0.30 \pm 0.01$ & $0.30 \pm 0.01$ & $0.30 \pm 0.01$ & $0.28 \pm 0.01$ & $0.24 \pm 0.01$ & $0.28 \pm 0.01$ & $0.27 \pm 0.01$ \\
P$\downarrow$ & $0.02 \pm 0.03$ & $0.06 \pm 0.01$ & $0.01 \pm 0.01$ & $0.03 \pm 0.02$ & $0.03 \pm 0.02$ & $0.01 \pm 0.01$ & $0.19 \pm 0.03$ & $0.00 \pm 0.00$ & $0.00 \pm 0.00$ & $0.19 \pm 0.01$ \\ \midrule
\multicolumn{11}{c}{\textbf{Concept: Chubby}} \\ \midrule
R$\uparrow$ & $1.67 \pm 1.60$ & $4.10 \pm 1.44$ & $3.89 \pm 1.61$ & $6.67 \pm 1.68$ & $3.64 \pm 1.83$ & $1.43 \pm 1.19$ & $-0.93 \pm 0.05$ & $-0.39 \pm 0.19$ & $0.67 \pm 1.45$ & $0.78 \pm 1.46$ \\
S$\downarrow$ & $0.29 \pm 0.01$ & $0.29 \pm 0.01$ & $0.28 \pm 0.01$ & $0.28 \pm 0.01$ & $0.29 \pm 0.01$ & $0.30 \pm 0.01$ & $0.32 \pm 0.01$ & $0.31 \pm 0.01$ & $0.30 \pm 0.01$ & $0.30 \pm 0.01$ \\
P$\downarrow$ & $0.02 \pm 0.02$ & $0.07 \pm 0.02$ & $0.01 \pm 0.01$ & $0.03 \pm 0.02$ & $0.03 \pm 0.02$ & $0.01 \pm 0.01$ & $0.17 \pm 0.03$ & $0.03 \pm 0.01$ & $0.20 \pm 0.01$ & $0.19 \pm 0.01$ \\ \midrule
\multicolumn{11}{c}{\textbf{Concept: Cluttered Room}} \\ \midrule
R$\uparrow$ & $2.01 \pm 0.72$ & $2.34 \pm 0.77$ & $1.14 \pm 0.86$ & $0.14 \pm 0.21$ & $1.19 \pm 1.05$ & $0.36 \pm 0.50$ & $-0.35 \pm 0.09$ & $-0.31 \pm 0.12$ & $0.81 \pm 0.84$ & $2.13 \pm 0.46$ \\
S$\downarrow$ & $0.30 \pm 0.01$ & $0.28 \pm 0.01$ & $0.28 \pm 0.01$ & $0.27 \pm 0.01$ & $0.29 \pm 0.01$ & $0.30 \pm 0.01$ & $0.30 \pm 0.01$ & $0.30 \pm 0.01$ & $0.28 \pm 0.01$ & $0.29 \pm 0.00$ \\
P$\downarrow$ & $0.10 \pm 0.04$ & $0.17 \pm 0.02$ & $0.03 \pm 0.02$ & $0.01 \pm 0.00$ & $0.07 \pm 0.03$ & $0.05 \pm 0.03$ & $0.22 \pm 0.03$ & $0.23 \pm 0.04$ & $0.09 \pm 0.01$ & $0.19 \pm 0.02$ \\ \midrule
\multicolumn{11}{c}{\textbf{Concept: Cooked Food}} \\ \midrule
R$\uparrow$ & $0.47 \pm 0.62$ & $0.00 \pm 0.00$ & $0.57 \pm 0.38$ & $0.11 \pm 0.18$ & $0.65 \pm 0.66$ & $0.00 \pm 0.00$ & $0.17 \pm 0.07$ & $0.00 \pm 0.00$ & $0.23 \pm 0.34$ & $0.24 \pm 0.14$ \\
S$\downarrow$ & $0.28 \pm 0.00$ & $0.28 \pm 0.00$ & $0.28 \pm 0.00$ & $0.28 \pm 0.00$ & $0.29 \pm 0.00$ & $0.29 \pm 0.00$ & $0.28 \pm 0.02$ & $0.29 \pm 0.02$ & $0.30 \pm 0.01$ & $0.29 \pm 0.00$ \\
P$\downarrow$ & $0.03 \pm 0.02$ & --- & $0.01 \pm 0.00$ & $0.00 \pm 0.00$ & $0.03 \pm 0.01$ & --- & $0.23 \pm 0.03$ & --- & $0.05 \pm 0.01$ & $0.10 \pm 0.01$ \\ \midrule
\multicolumn{11}{c}{\textbf{Concept: Damaged Car}} \\ \midrule
R$\uparrow$ & $0.65 \pm 0.94$ & $1.37 \pm 0.96$ & $1.04 \pm 0.82$ & $0.43 \pm 0.35$ & $0.01 \pm 0.74$ & $0.14 \pm 0.44$ & $0.25 \pm 0.04$ & $0.02 \pm 0.22$ & $1.15 \pm 1.09$ & $1.34 \pm 1.17$ \\
S$\downarrow$ & $0.28 \pm 0.00$ & $0.28 \pm 0.00$ & $0.29 \pm 0.01$ & $0.30 \pm 0.00$ & $0.29 \pm 0.01$ & $0.29 \pm 0.01$ & $0.28 \pm 0.01$ & $0.29 \pm 0.01$ & $0.28 \pm 0.01$ & $0.29 \pm 0.01$ \\
P$\downarrow$ & $0.04 \pm 0.03$ & $0.08 \pm 0.01$ & $0.01 \pm 0.01$ & $0.01 \pm 0.01$ & $0.04 \pm 0.02$ & $0.01 \pm 0.01$ & $0.17 \pm 0.01$ & $0.19 \pm 0.03$ & $0.13 \pm 0.02$ & $0.08 \pm 0.01$ \\ \midrule
\multicolumn{11}{c}{\textbf{Concept: Glasses}} \\ \midrule
R$\uparrow$ & $3.46 \pm 0.85$ & $3.81 \pm 0.61$ & $3.30 \pm 0.88$ & $3.48 \pm 0.93$ & $3.26 \pm 1.00$ & $2.86 \pm 1.27$ & $0.56 \pm 0.26$ & $0.47 \pm 0.15$ & $0.00 \pm 0.00$ & $0.00 \pm 0.00$ \\
S$\downarrow$ & $0.27 \pm 0.01$ & $0.27 \pm 0.01$ & $0.28 \pm 0.01$ & $0.28 \pm 0.01$ & $0.28 \pm 0.01$ & $0.20 \pm 0.01$ & $0.32 \pm 0.01$ & $0.29 \pm 0.01$ & $0.28 \pm 0.01$ & $0.23 \pm 0.01$ \\
P$\downarrow$ & $0.03 \pm 0.03$ & $0.04 \pm 0.00$ & $0.01 \pm 0.01$ & $0.02 \pm 0.01$ & $0.02 \pm 0.01$ & $0.02 \pm 0.01$ & $0.05 \pm 0.03$ & $0.03 \pm 0.01$ & --- & --- \\ \midrule
\multicolumn{11}{c}{\textbf{Concept: Lipstick}} \\ \midrule
R$\uparrow$ & $1.22 \pm 0.92$ & $0.11 \pm 0.27$ & $2.42 \pm 1.08$ & $2.26 \pm 1.05$ & $2.28 \pm 1.22$ & $0.40 \pm 0.67$ & $-2.75 \pm 0.81$ & $-0.00 \pm 0.04$ & $0.00 \pm 0.00$ & $0.00 \pm 0.00$ \\
S$\downarrow$ & $0.27 \pm 0.01$ & $0.29 \pm 0.01$ & $0.27 \pm 0.01$ & $0.27 \pm 0.01$ & $0.27 \pm 0.01$ & $0.28 \pm 0.01$ & $0.29 \pm 0.01$ & $0.29 \pm 0.01$ & $0.30 \pm 0.01$ & $0.21 \pm 0.01$ \\
P$\downarrow$ & $0.02 \pm 0.02$ & $0.00 \pm 0.00$ & $0.01 \pm 0.01$ & $0.01 \pm 0.01$ & $0.05 \pm 0.03$ & $0.01 \pm 0.01$ & $0.13 \pm 0.03$ & $0.00 \pm 0.00$ & $0.00 \pm 0.00$ & $0.06 \pm 0.01$ \\ \midrule
\multicolumn{11}{c}{\textbf{Concept: Surprised}} \\ \midrule
R$\uparrow$ & $0.23 \pm 0.43$ & $0.41 \pm 0.55$ & $0.46 \pm 0.44$ & $0.82 \pm 0.54$ & $-0.20 \pm 0.51$ & $0.54 \pm 0.63$ & $-0.86 \pm 0.04$ & $0.04 \pm 0.22$ & $0.00 \pm 0.00$ & $0.00 \pm 0.00$ \\
S$\downarrow$ & $0.30 \pm 0.01$ & $0.30 \pm 0.01$ & $0.29 \pm 0.01$ & $0.28 \pm 0.01$ & $0.29 \pm 0.01$ & $0.29 \pm 0.01$ & $0.30 \pm 0.01$ & $0.31 \pm 0.01$ & $0.31 \pm 0.00$ & $0.25 \pm 0.01$ \\
P$\downarrow$ & $0.01 \pm 0.01$ & $0.06 \pm 0.01$ & $0.01 \pm 0.01$ & $0.02 \pm 0.01$ & $0.03 \pm 0.02$ & $0.01 \pm 0.01$ & $0.37 \pm 0.02$ & $0.03 \pm 0.01$ & $0.06 \pm 0.01$ & $0.22 \pm 0.02$ \\ \midrule
\multicolumn{11}{c}{\textbf{Concept: Smiling}} \\ \midrule
R$\uparrow$ & $2.30 \pm 1.86$ & $5.05 \pm 1.22$ & $4.54 \pm 1.41$ & $5.87 \pm 1.28$ & $2.95 \pm 1.60$ & $1.45 \pm 1.46$ & $-4.70 \pm 0.47$ & $0.85 \pm 0.62$ & $0.99 \pm 1.13$ & $1.17 \pm 1.63$ \\
S$\downarrow$ & $0.29 \pm 0.01$ & $0.29 \pm 0.01$ & $0.28 \pm 0.01$ & $0.28 \pm 0.01$ & $0.30 \pm 0.01$ & $0.31 \pm 0.01$ & $0.29 \pm 0.01$ & $0.31 \pm 0.02$ & $0.30 \pm 0.01$ & $0.29 \pm 0.01$ \\
P$\downarrow$ & $0.02 \pm 0.02$ & $0.02 \pm 0.01$ & $0.00 \pm 0.01$ & $0.01 \pm 0.02$ & $0.02 \pm 0.01$ & $0.06 \pm 0.04$ & $0.20 \pm 0.01$ & $0.04 \pm 0.01$ & $0.06 \pm 0.01$ & $0.06 \pm 0.01$ \\ \midrule
\multicolumn{11}{c}{\textbf{Overall Score (Averaged across concepts)}} \\  \midrule
R$\uparrow$ & $1.60 \pm 1.08$ & $2.54 \pm 0.78$ & $2.34 \pm 1.05$ & $2.85 \pm 0.86$ & $1.73 \pm 1.13$ & $0.93 \pm 0.80$ & $-0.92 \pm 0.74$ & $0.08 \pm 0.37$ & $0.38 \pm 0.48$ & $0.57 \pm 0.55$ \\
S$\downarrow$ & $0.29 \pm 0.01$ & $0.28 \pm 0.01$ & $0.28 \pm 0.01$ & $0.28 \pm 0.01$ & $0.29 \pm 0.01$ & $0.28 \pm 0.01$ & $0.30 \pm 0.02$ & $0.29 \pm 0.01$ & $0.29 \pm 0.01$ & $0.27 \pm 0.01$ \\
P$\downarrow$ & $0.03 \pm 0.02$ & $0.06 \pm 0.03$ & $0.01 \pm 0.01$ & $0.02 \pm 0.01$ & $0.03 \pm 0.02$ & $0.02 \pm 0.01$ & $0.18 \pm 0.04$ & $0.06 \pm 0.03$ & $0.06 \pm 0.03$ & $0.13 \pm 0.01$ \\ \bottomrule
\end{tabular}}
\end{table*}

\begin{table*}[!ht]
\centering
\caption{Comparison of video results with and without ASTD. Metrics include Range (R), Smoothness (S), and Preservation (P), each measured for Static (S) and Dynamic (D) aspects.}
\label{tab:all_video_results}
\resizebox{1\textwidth}{!}{%
\begin{tabular}{@{}ll|cc|cc|cc|cc@{}}
\toprule
\multicolumn{2}{c|}{\textbf{Concept}} & \multicolumn{2}{c|}{\textbf{CS}} & \multicolumn{2}{c|}{\textbf{TE}} & \multicolumn{2}{c|}{\textbf{Ours}} & \multicolumn{2}{c}{\textbf{PromptSliders}} \\
\cmidrule(lr){3-4} \cmidrule(lr){5-6} \cmidrule(lr){7-8} \cmidrule(lr){9-10}
& \textbf{Metric} & \textbf{w/o ASTD} & \textbf{w/ ASTD} & \textbf{w/o ASTD} & \textbf{w/ ASTD} & \textbf{w/o ASTD} & \textbf{w/ ASTD} & \textbf{w/o ASTD} & \textbf{w/ ASTD} \\ \midrule
\multirow{6}{*}{\textbf{Mountain Hiking}} 
& R(S) $\uparrow$ & $2.58 \pm 0.43$ & $0.81 \pm 0.46$ & $0.94 \pm 0.25$ & $1.19 \pm 0.53$ & $1.08 \pm 0.44$ & $1.24 \pm 0.52$ & $-1.24 \pm 0.01$ & $-1.22 \pm 0.01$ \\
& R(D) $\uparrow$ & $0.04 \pm 0.00$ & $0.01 \pm 0.01$ & $0.02 \pm 0.00$ & $0.01 \pm 0.00$ & $0.01 \pm 0.01$ & $0.02 \pm 0.01$ & $-0.01 \pm 0.01$ &  $-0.01 \pm 0.01$ \\
& P(S) $\downarrow$ & $0.10 \pm 0.02$ & $0.08 \pm 0.02$ & $0.06 \pm 0.02$ & $0.06 \pm 0.03$ & $0.03 \pm 0.01$ & $0.03 \pm 0.01$ & $0.09 \pm 0.01$ & $0.08 \pm 0.01$\\
& P(D) $\downarrow$ & $-0.54 \pm 0.26$ & $-0.44 \pm 0.20$ & $-0.39 \pm 0.27$ & $-0.46 \pm 0.31$ & $-0.28 \pm 0.16$ & $-0.28 \pm 0.16$ & $-1.02 \pm 0.01$ & $-1.00 \pm 0.01$\\
& S(S) $\downarrow$ & $0.33 \pm 0.01$ & $0.34 \pm 0.01$ & $0.34 \pm 0.01$ & $0.33 \pm 0.01$ & $0.34 \pm 0.01$ & $0.34 \pm 0.01$ & $0.35 \pm 0.01$ & $0.36 \pm 0.01$\\
& S(D) $\uparrow$ & $0.44 \pm 0.02$ & $0.44 \pm 0.01$ & $0.43 \pm 0.02$ & $0.44 \pm 0.01$ & $0.44 \pm 0.02$ & $0.43 \pm 0.02$ & $0.41 \pm 0.01$ & $0.43 \pm 0.01$ \\ \midrule
\multirow{6}{*}{\textbf{Sailboat}} 
& R(S) $\uparrow$ & $4.05 \pm 0.35$ & $4.98 \pm 0.79$ & $0.21 \pm 0.37$ & $0.68 \pm 0.43$ & $0.90 \pm 0.34$ & $2.05 \pm 0.44$ & $0.07 \pm 0.01$ & $-0.28 \pm 0.01$\\
& R(D) $\uparrow$ & $0.06 \pm 0.01$ & $0.08 \pm 0.02$ & $0.01 \pm 0.00$ & $0.01 \pm 0.01$ & $0.01 \pm 0.00$ & $0.03 \pm 0.01$ & $-0.01 \pm 0.01$ & $-0.01 \pm 0.01$\\
& P(S) $\downarrow$ & $0.08 \pm 0.03$ & $0.11 \pm 0.03$ & $0.02 \pm 0.00$ & $0.02 \pm 0.00$ & $0.02 \pm 0.00$ & $0.03 \pm 0.01$ & $0.09 \pm 0.01$ & $0.08 \pm 0.01$\\
& P(D) $\downarrow$ & $-3.54 \pm 1.54$ & $-3.89 \pm 1.56$ & $-1.40 \pm 0.49$ & $-1.35 \pm 0.46$ & $-1.19 \pm 0.47$ & $-1.56 \pm 0.62$ & $-1.51 \pm 0.01$ & $-1.53 \pm 0.01$\\
& S(S) $\downarrow$ & $0.29 \pm 0.01$ & $0.28 \pm 0.00$ & $0.32 \pm 0.01$ & $0.32 \pm 0.01$ & $0.31 \pm 0.01$ & $0.31 \pm 0.01$ & $0.33 \pm 0.01$ & $0.33 \pm 0.01$\\
& S(D) $\uparrow$ & $0.41 \pm 0.01$ & $0.41 \pm 0.01$ & $0.45 \pm 0.02$ & $0.44 \pm 0.02$ & $0.43 \pm 0.01$ & $0.44 \pm 0.02$ & $0.43 \pm 0.01$ & $0.47 \pm 0.01$\\ \midrule
\multirow{6}{*}{\textbf{Car Type}} 
& R(S) $\uparrow$ & $-0.17 \pm 1.09$ & $2.08 \pm 0.89$ & $4.55 \pm 0.68$ & $4.42 \pm 1.01$ & $2.97 \pm 0.80$ & $2.93 \pm 0.78$ & $-0.21 \pm 0.01$ & $-0.28 \pm 0.01$\\
& R(D) $\uparrow$ & $0.00 \pm 0.01$ & $0.02 \pm 0.01$ & $0.06 \pm 0.01$ & $0.06 \pm 0.02$ & $0.05 \pm 0.02$ & $0.05 \pm 0.02$ & $0.01 \pm 0.01$ & $0.01 \pm 0.01$\\
& P(S) $\downarrow$ & $0.10 \pm 0.03$ & $0.14 \pm 0.03$ & $0.07 \pm 0.02$ & $0.07 \pm 0.02$ & $0.05 \pm 0.01$ & $0.05 \pm 0.01$ & $0.07 \pm 0.01$ & $0.07 \pm 0.01$\\
& P(D) $\downarrow$ & $-5.11 \pm 3.41$ & $-5.55 \pm 3.72$ & $-4.80 \pm 2.28$ & $-4.82 \pm 2.22$ & $-3.76 \pm 1.75$ & $-3.80 \pm 1.82$ & $-1.01 \pm 0.01$ & $-1.09 \pm 0.01$\\
& S(S) $\downarrow$ & $0.36 \pm 0.01$ & $0.32 \pm 0.01$ & $0.32 \pm 0.01$ & $0.31 \pm 0.01$ & $0.33 \pm 0.01$ & $0.32 \pm 0.01$ & $0.38 \pm 0.01$ & $0.39 \pm 0.01$\\
& S(D) $\uparrow$ & $0.46 \pm 0.02$ & $0.44 \pm 0.02$ & $0.43 \pm 0.02$ & $0.41 \pm 0.01$ & $0.43 \pm 0.01$ & $0.42 \pm 0.01$ & $0.45 \pm 0.01$ & $0.45 \pm 0.01$\\ \midrule
\multirow{6}{*}{\textbf{River Waves}} 
& R(S) $\uparrow$ & $1.88 \pm 0.89$ & $0.79 \pm 0.51$ & $1.26 \pm 0.69$ & $0.22 \pm 0.31$ & $0.66 \pm 0.65$ & $0.61 \pm 0.30$ & $-0.53 \pm 0.01$ & $1.23 \pm 0.01$\\
& R(D) $\uparrow$ & $0.03 \pm 0.02$ & $0.01 \pm 0.01$ & $0.01 \pm 0.01$ & $0.00 \pm 0.01$ & $0.01 \pm 0.00$ & $0.01 \pm 0.01$ & $-0.02 \pm 0.01$ & $0.01 \pm 0.01$\\
& P(S) $\downarrow$ & $0.09 \pm 0.03$ & $0.06 \pm 0.02$ & $0.04 \pm 0.01$ & $0.03 \pm 0.01$ & $0.03 \pm 0.01$ & $0.03 \pm 0.01$ & $0.14 \pm 0.01$ & $0.10 \pm 0.01$\\
& P(D) $\downarrow$ & $-0.39 \pm 0.23$ & $-0.30 \pm 0.17$ & $-0.23 \pm 0.15$ & $-0.22 \pm 0.14$ & $-0.23 \pm 0.12$ & $-0.22 \pm 0.12$ & $-0.36 \pm 0.01$ & $-0.35 \pm 0.01$\\
& S(S) $\downarrow$ & $0.32 \pm 0.01$ & $0.33 \pm 0.01$ & $0.32 \pm 0.01$ & $0.33 \pm 0.01$ & $0.32 \pm 0.01$ & $0.32 \pm 0.01$ & $0.32 \pm 0.01$ & $0.31 \pm 0.01$\\
& S(D) $\uparrow$ & $0.46 \pm 0.02$ & $0.47 \pm 0.03$ & $0.44 \pm 0.02$ & $0.45 \pm 0.02$ & $0.44 \pm 0.01$ & $0.45 \pm 0.01$ & $0.43 \pm 0.01$ & $0.45 \pm 0.01$\\ \midrule
\multirow{6}{*}{\textbf{Butterfly Wings}} 
& R(S) $\uparrow$ & $0.16 \pm 0.24$ & $0.06 \pm 0.11$ & $0.04 \pm 0.31$ & $0.20 \pm 0.30$ & $0.14 \pm 0.14$ & $0.01 \pm 0.12$ & $-0.01 \pm 0.01$ & $-0.23 \pm 0.01$\\
& R(D) $\uparrow$ & $0.00 \pm 0.00$ & $0.00 \pm 0.01$ & $0.00 \pm 0.00$ & $-0.00 \pm 0.01$ & $0.00 \pm 0.00$ & $0.00 \pm 0.00$ & $-0.01 \pm 0.01$ & $-0.01 \pm 0.01$\\
& P(S) $\downarrow$ & $0.04 \pm 0.01$ & $0.04 \pm 0.01$ & $0.05 \pm 0.02$ & $0.05 \pm 0.02$ & $0.04 \pm 0.02$ & $0.04 \pm 0.02$ & $0.21 \pm 0.01$ & $0.20 \pm 0.01$\\
& P(D) $\downarrow$ & $-0.82 \pm 0.33$ & $-0.80 \pm 0.32$ & $-1.05 \pm 0.41$ & $-1.06 \pm 0.36$ & $-0.82 \pm 0.32$ & $-0.83 \pm 0.32$ & $-3.15 \pm 0.01$ & $ -3.14\pm 0.01$\\
& S(S) $\downarrow$ & $0.30 \pm 0.01$ & $0.30 \pm 0.01$ & $0.30 \pm 0.01$ & $0.30 \pm 0.00$ & $0.30 \pm 0.00$ & $0.30 \pm 0.00$ & $0.30 \pm 0.01$ & $0.30 \pm 0.01$\\
& S(D) $\uparrow$ & $0.45 \pm 0.02$ & $0.46 \pm 0.02$ & $0.45 \pm 0.02$ & $0.46 \pm 0.02$ & $0.45 \pm 0.02$ & $0.45 \pm 0.01$ & $0.45 \pm 0.01$ & $0.47 \pm 0.01$\\ \midrule
\multirow{6}{*}{\textbf{Mountain Type}} 
& R(S) $\uparrow$ & $2.46 \pm 0.91$ & $0.38 \pm 0.44$ & $2.46 \pm 1.10$ & $1.91 \pm 0.87$ & $1.59 \pm 0.63$ & $1.33 \pm 0.80$ & $1.52 \pm 0.01$ & $-0.52 \pm 0.01$\\
& R(D) $\uparrow$ & $0.02 \pm 0.01$ & $0.00 \pm 0.01$ & $0.04 \pm 0.01$ & $0.03 \pm 0.01$ & $0.03 \pm 0.01$ & $0.02 \pm 0.01$ & $0.02 \pm 0.01$ & $-0.01 \pm 0.01$\\
& P(S) $\downarrow$ & $0.08 \pm 0.04$ & $0.06 \pm 0.03$ & $0.05 \pm 0.03$ & $0.05 \pm 0.03$ & $0.03 \pm 0.01$ & $0.03 \pm 0.01$ & $0.20 \pm 0.01$ & $0.19 \pm 0.01$\\
& P(D) $\downarrow$ & $-0.89 \pm 1.46$ & $-0.70 \pm 1.14$ & $-0.81 \pm 1.46$ & $-0.78 \pm 1.39$ & $-0.69 \pm 1.23$ & $-0.64 \pm 1.11$ & $-0.50 \pm 0.01$ & $-0.48 \pm 0.01$\\
& S(S) $\downarrow$& $0.33 \pm 0.01$ & $0.32 \pm 0.01$ & $0.31 \pm 0.01$ & $0.32 \pm 0.01$ & $0.31 \pm 0.01$ & $0.32 \pm 0.01$ & $0.31 \pm 0.01$ & $0.32 \pm 0.01$\\
& S(D) $\uparrow$ & $0.47 \pm 0.02$ & $0.48 \pm 0.01$ & $0.45 \pm 0.02$ & $0.45 \pm 0.02$ & $0.43 \pm 0.02$ & $0.43 \pm 0.02$ & $0.45 \pm 0.01$ & $0.43 \pm 0.01$\\ \midrule
\multirow{6}{*}{\textbf{Tree}} 
& R(S) $\uparrow$ & $0.59 \pm 0.61$ & $3.01 \pm 0.00$ & $0.47 \pm 1.11$ & $0.15 \pm 0.38$ & $1.05 \pm 1.02$ & $2.34 \pm 1.19$ & $-0.98 \pm 0.01$ & $-0.51 \pm 0.01$ \\
& R(D) $\uparrow$ & $0.02 \pm 0.02$ & $0.00 \pm 0.00$ & $0.02 \pm 0.01$ & $0.00 \pm 0.01$ & $0.02 \pm 0.01$ & $0.05 \pm 0.01$ & $-0.03 \pm 0.01$ & $-0.01 \pm 0.01$\\
& P(S) $\downarrow$ & $0.09 \pm 0.04$ & $0.09 \pm 0.04$ & $0.03 \pm 0.02$ & $0.03 \pm 0.01$ & $0.03 \pm 0.01$ & $0.05 \pm 0.02$ & $0.22 \pm 0.01$ & $0.19 \pm 0.01$\\
& P(D) $\downarrow$ & $-0.57 \pm 0.41$ & $-0.57 \pm 0.41$ & $-0.55 \pm 0.41$ & $-0.46 \pm 0.34$ & $-0.41 \pm 0.24$ & $-0.52 \pm 0.37$ & $-0.50 \pm 0.01$ & $-0.48 \pm 0.01$\\
& S(S) $\downarrow$ & $0.32 \pm 0.02$ & $0.27 \pm 0.01$ & $0.31 \pm 0.01$ & $0.32 \pm 0.01$ & $0.32 \pm 0.01$ & $0.31 \pm 0.01$ & $0.31 \pm 0.01$ & $0.32 \pm 0.01$\\
& S(D) $\uparrow$ & $0.47 \pm 0.02$ & $0.50 \pm 0.00$ & $0.44 \pm 0.02$ & $0.45 \pm 0.02$ & $0.44 \pm 0.02$ & $0.43 \pm 0.01$ & $0.46 \pm 0.01$ & $0.43 \pm 0.01$\\ \midrule
\multirow{6}{*}{\textbf{Cat Fat}} 
& R(S) $\uparrow$ & $-0.27 \pm 0.49$ & $-0.60 \pm 0.70$ & $0.43 \pm 0.39$ & $0.00 \pm 0.00$ & $0.35 \pm 0.50$ & $0.21 \pm 0.34$ & $1.76 \pm 0.01$ & $0.15 \pm 0.01$\\
& R(D) $\uparrow$ & $-0.01 \pm 0.01$ & $-0.00 \pm 0.01$ & $0.01 \pm 0.01$ & $0.00 \pm 0.00$ & $0.01 \pm 0.01$ & $0.00 \pm 0.01$ & $0.01 \pm 0.01$ & $-0.02 \pm 0.01$\\
& P(S) $\downarrow$ & $0.06 \pm 0.03$ & $0.09 \pm 0.02$ & $0.03 \pm 0.01$ & $0.03 \pm 0.01$ & $0.02 \pm 0.01$ & $0.02 \pm 0.01$ & $0.11 \pm 0.01$ & $0.09 \pm 0.01$\\
& P(D) $\downarrow$ & $-0.70 \pm 0.26$ & $-0.87 \pm 0.36$ & $-0.60 \pm 0.24$ & $-0.60 \pm 0.24$ & $-0.45 \pm 0.18$ & $-0.46 \pm 0.17$ & $-0.80 \pm 0.01$ & $-0.71 \pm 0.01$\\
& S(S) $\downarrow$ & $0.30 \pm 0.02$ & $0.30 \pm 0.00$ & $0.30 \pm 0.01$ & $0.25 \pm 0.02$ & $0.30 \pm 0.01$ & $0.31 \pm 0.01$ & $0.30 \pm 0.01$ & $0.30 \pm 0.01$\\
& S(D) $\uparrow$ & $0.47 \pm 0.02$ & $0.47 \pm 0.02$ & $0.45 \pm 0.02$ & $0.50 \pm 0.00$ & $0.43 \pm 0.02$ & $0.45 \pm 0.01$ & $0.47 \pm 0.01$ & $0.47 \pm 0.01$\\ \midrule
\multirow{6}{*}{\textbf{Silhouette}} 
& R(S) $\uparrow$ & $0.19 \pm 0.27$ & $0.18 \pm 0.23$ & $2.61 \pm 0.98$ & $2.46 \pm 1.05$ & $1.19 \pm 0.92$ & $2.85 \pm 0.94$ & $-0.04 \pm 0.01$ & $-0.02 \pm 0.01$\\
& R(D) $\uparrow$ & $0.01 \pm 0.01$ & $0.00 \pm 0.01$ & $0.04 \pm 0.01$ & $0.04 \pm 0.02$ & $0.02 \pm 0.02$ & $0.04 \pm 0.02$ & $-0.01 \pm 0.01$ & $-0.01 \pm 0.01$\\
& P(S) $\downarrow$ & $0.04 \pm 0.02$ & $0.05 \pm 0.01$ & $0.02 \pm 0.00$ & $0.03 \pm 0.00$ & $0.02 \pm 0.00$ & $0.02 \pm 0.01$ & $0.16 \pm 0.01$ & $0.17 \pm 0.01$\\
& P(D) $\downarrow$ & $-0.98 \pm 0.86$ & $-1.14 \pm 1.06$ & $-0.91 \pm 0.62$ & $-0.99 \pm 0.66$ & $-0.55 \pm 0.30$ & $-0.65 \pm 0.37$ & $-0.49 \pm 0.01$ & $-0.51 \pm 0.01$ \\
& S(S) $\downarrow$ & $0.30 \pm 0.01$ & $0.30 \pm 0.01$ & $0.29 \pm 0.01$ & $0.29 \pm 0.01$ & $0.30 \pm 0.01$ & $0.29 \pm 0.01$ & $0.30 \pm 0.01$ & $0.31 \pm 0.01$\\
& S(D) $\uparrow$ & $0.45 \pm 0.02$ & $0.46 \pm 0.02$ & $0.42 \pm 0.02$ & $0.44 \pm 0.01$ & $0.44 \pm 0.02$ & $0.43 \pm 0.01$ & $0.44 \pm 0.01$ & $0.45 \pm 0.01$\\ \midrule
\multirow{6}{*}{\textbf{Lighthouse Water}} 
& R(S) $\uparrow$ & $0.28 \pm 0.56$ & $0.01 \pm 0.33$ & $0.85 \pm 0.40$ & $1.73 \pm 0.68$ & $0.68 \pm 0.29$ & $1.62 \pm 0.59$ & $0.04 \pm 0.01$ & $-0.04 \pm 0.01$\\
& R(D) $\uparrow$ & $0.01 \pm 0.01$ & $-0.00 \pm 0.01$ & $0.01 \pm 0.00$ & $0.02 \pm 0.01$ & $0.01 \pm 0.01$ & $0.02 \pm 0.01$ & $0.00 \pm 0.01$ & $0.02 \pm 0.01$ \\
& P(S) $\downarrow$ & $0.08 \pm 0.03$ & $0.06 \pm 0.02$ & $0.03 \pm 0.01$ & $0.04 \pm 0.01$ & $0.02 \pm 0.01$ & $0.03 \pm 0.01$ & $0.03 \pm 0.01$ & $0.02 \pm 0.01$\\
& P(D) $\downarrow$ & $-1.50 \pm 0.53$ & $-1.30 \pm 0.48$ & $-0.99 \pm 0.46$ & $-1.11 \pm 0.51$ & $-0.75 \pm 0.31$ & $-0.98 \pm 0.39$ & $-0.43 \pm 0.01$ & $-0.40 \pm 0.01$\\
& S(S) $\downarrow$ & $0.32 \pm 0.01$ & $0.30 \pm 0.01$ & $0.29 \pm 0.00$ & $0.30 \pm 0.01$ & $0.30 \pm 0.01$ & $0.30 \pm 0.01$ & $0.31 \pm 0.01$ & $0.32 \pm 0.01$\\
& S(D) $\uparrow$ & $0.47 \pm 0.02$ & $0.46 \pm 0.02$ & $0.44 \pm 0.02$ & $0.44 \pm 0.01$ & $0.44 \pm 0.02$ & $0.43 \pm 0.02$ & $0.43 \pm 0.01$ & $0.50 \pm 0.01$\\ \midrule

\multirow{6}{*}{\textbf{Overall Avg}} 

& R(S) $\uparrow$ & $1.17 \pm 0.58$ & $2.29 \pm 0.45$ & $1.38 \pm 0.61$ & $1.30 \pm 0.56$ & $1.06 \pm 0.57$ & $1.52 \pm 0.60$ & $0.04 \pm 0.01$ & $-0.17 \pm 0.01$ \\

& R(D) $\uparrow$ & $0.02 \pm 0.01$ & $0.02 \pm 0.01$ & $0.02 \pm 0.01$ & $0.02 \pm 0.01$ & $0.02 \pm 0.01$ & $0.03 \pm 0.01$ & $-0.01 \pm 0.01$ & $-0.01 \pm 0.01$ \\

& P(S) $\downarrow$ & $0.08 \pm 0.02$ & $0.08 \pm 0.02$ & $0.04 \pm 0.01$ & $0.04 \pm 0.01$ & $0.03 \pm 0.01$ & $0.03 \pm 0.01$ & $0.13 \pm 0.01$ & $0.11 \pm 0.01$ \\

& P(D) $\downarrow$ & $-1.50 \pm 0.93$ & $-1.55 \pm 0.94$ & $-1.18 \pm 0.68$ & $-1.19 \pm 0.66$ & $-0.91 \pm 0.50$ & $-0.99 \pm 0.54$ & $-0.93 \pm 0.01$ & $-0.93 \pm 0.01$ \\

& S(S) $\downarrow$ & $0.32 \pm 0.01$ & $0.31 \pm 0.01$ & $0.31 \pm 0.01$ & $0.31 \pm 0.01$ & $0.32 \pm 0.01$ & $0.32 \pm 0.01$ & $0.32 \pm 0.01$ & $0.33 \pm 0.01$ \\

& S(D) $\uparrow$ & $0.45 \pm 0.02$ & $0.46 \pm 0.02$ & $0.44 \pm 0.02$ & $0.44 \pm 0.01$ & $0.44 \pm 0.02$ & $0.44 \pm 0.01$ & $0.44 \pm 0.01$ & $0.46 \pm 0.01$ \\ 
\bottomrule
\end{tabular}%
}
\end{table*}

\begin{table*}[!ht]
\centering
\caption{Overall results for audio evaluation. R=Conceptual Range, S=Smoothness, P=Preservation.}
\label{tab:all_audio_results}
\resizebox{1.0\textwidth}{!}{
\begin{tabular}{l|cc|cc|cc|cc}
\toprule
 & \multicolumn{2}{c|}{\textbf{CS}} & \multicolumn{2}{c|}{\textbf{Ours}} & \multicolumn{2}{c|}{\textbf{TE}} & \multicolumn{2}{c}{\textbf{P2P}} \\
\cmidrule(lr){2-3} \cmidrule(lr){4-5} \cmidrule(lr){6-7} \cmidrule(lr){8-9}
 & \textbf{w/o ASTD} & \textbf{w/ ASTD} & \textbf{w/o ASTD} & \textbf{w/ ASTD} & \textbf{w/o ASTD} & \textbf{w/ ASTD} & \textbf{w/o ASTD} & \textbf{w/ ASTD} \\ \midrule

\multicolumn{9}{c}{\textbf{Concept: Barking}} \\ \midrule
R$\uparrow$ & $1.19 \pm 1.72$ & $1.21 \pm 0.96$ & $0.52 \pm 1.24$ & $2.03 \pm 1.92$ & $4.52 \pm 1.87$ & $4.52 \pm 1.87$ & $6.88 \pm 1.99$ & $8.61 \pm 3.07$ \\
S$\downarrow$ & $0.34 \pm 0.03$ & $0.34 \pm 0.02$ & $0.34 \pm 0.03$ & $0.35 \pm 0.03$ & $0.34 \pm 0.02$ & $0.34 \pm 0.02$ & $0.36 \pm 0.03$ & $0.36 \pm 0.03$ \\
P$\downarrow$ & $0.89 \pm 0.45$ & $0.72 \pm 0.33$ & $0.76 \pm 0.45$ & $1.04 \pm 0.32$ & $2.72 \pm 0.46$ & $2.72 \pm 0.46$ & $3.39 \pm 0.82$ & $3.50 \pm 0.73$ \\ \midrule
\multicolumn{9}{c}{\textbf{Concept: Car}} \\ \midrule
R$\uparrow$ & $-6.48 \pm 4.61$ & $-2.32 \pm 1.35$ & $4.76 \pm 4.46$ & $7.06 \pm 4.88$ & $11.66 \pm 1.92$ & $3.27 \pm 2.12$ & $3.89 \pm 2.86$ & $1.31 \pm 3.22$ \\
S$\downarrow$ & $0.42 \pm 0.02$ & $0.35 \pm 0.01$ & $0.38 \pm 0.04$ & $0.36 \pm 0.04$ & $0.39 \pm 0.02$ & $0.40 \pm 0.02$ & $0.39 \pm 0.03$ & $0.39 \pm 0.02$ \\
P$\downarrow$ & $3.39 \pm 0.79$ & $5.55 \pm 0.35$ & $1.76 \pm 0.93$ & $2.19 \pm 0.72$ & $5.49 \pm 0.23$ & $3.40 \pm 0.41$ & $3.34 \pm 1.00$ & $3.94 \pm 0.75$ \\ \midrule
\multicolumn{9}{c}{\textbf{Concept: Cat}} \\ \midrule
R$\uparrow$ & $1.99 \pm 1.85$ & $6.66 \pm 1.83$ & $1.16 \pm 1.38$ & $5.59 \pm 2.73$ & $4.35 \pm 4.30$ & $12.30 \pm 3.58$ & $1.63 \pm 1.45$ & $1.63 \pm 1.45$ \\
S$\downarrow$ & $0.34 \pm 0.02$ & $0.34 \pm 0.02$ & $0.34 \pm 0.02$ & $0.34 \pm 0.04$ & $0.41 \pm 0.02$ & $0.44 \pm 0.03$ & $0.36 \pm 0.03$ & $0.36 \pm 0.03$ \\
P$\downarrow$ & $1.27 \pm 0.44$ & $2.75 \pm 0.43$ & $0.94 \pm 0.59$ & $2.44 \pm 0.69$ & $4.51 \pm 0.30$ & $5.64 \pm 0.42$ & $3.34 \pm 0.36$ & $3.34 \pm 0.36$ \\ \midrule
\multicolumn{9}{c}{\textbf{Concept: Choir}} \\ \midrule
R$\uparrow$ & $1.48 \pm 0.76$ & $6.91 \pm 0.40$ & $0.55 \pm 0.56$ & $2.84 \pm 0.82$ & $1.05 \pm 0.48$ & $2.72 \pm 0.53$ & $0.47 \pm 0.57$ & $0.26 \pm 0.80$ \\
S$\downarrow$ & $0.38 \pm 0.03$ & $0.40 \pm 0.01$ & $0.38 \pm 0.02$ & $0.38 \pm 0.02$ & $0.37 \pm 0.02$ & $0.45 \pm 0.01$ & $0.41 \pm 0.03$ & $0.41 \pm 0.02$ \\
P$\downarrow$ & $1.65 \pm 0.25$ & $3.61 \pm 0.38$ & $0.95 \pm 0.33$ & $2.45 \pm 0.32$ & $1.86 \pm 0.33$ & $5.44 \pm 0.15$ & $3.33 \pm 0.76$ & $3.39 \pm 0.72$ \\ \midrule
\multicolumn{9}{c}{\textbf{Concept: Clapping}} \\ \midrule
R$\uparrow$ & $0.20 \pm 0.27$ & $-0.05 \pm 0.66$ & $0.33 \pm 0.31$ & $1.42 \pm 0.57$ & $0.33 \pm 0.34$ & $0.45 \pm 0.84$ & $0.24 \pm 0.50$ & $-0.65 \pm 0.47$ \\
S$\downarrow$ & $0.36 \pm 0.02$ & $0.42 \pm 0.01$ & $0.35 \pm 0.02$ & $0.33 \pm 0.02$ & $0.46 \pm 0.01$ & $0.42 \pm 0.02$ & $0.37 \pm 0.03$ & $0.45 \pm 0.02$ \\
P$\downarrow$ & $0.66 \pm 0.34$ & $2.52 \pm 0.21$ & $0.41 \pm 0.19$ & $1.89 \pm 0.31$ & $4.99 \pm 0.21$ & $4.06 \pm 0.22$ & $2.51 \pm 0.76$ & $4.54 \pm 0.45$ \\ \midrule
\multicolumn{9}{c}{\textbf{Concept: Crowd}} \\ \midrule
R$\uparrow$ & $14.27 \pm 1.27$ & $13.74 \pm 1.39$ & $4.37 \pm 1.38$ & $9.34 \pm 2.20$ & $-1.58 \pm 0.77$ & $-1.61 \pm 0.77$ & $4.89 \pm 1.53$ & $12.29 \pm 2.21$ \\
S$\downarrow$ & $0.38 \pm 0.02$ & $0.37 \pm 0.01$ & $0.36 \pm 0.02$ & $0.36 \pm 0.02$ & $0.47 \pm 0.02$ & $0.46 \pm 0.02$ & $0.48 \pm 0.01$ & $0.41 \pm 0.02$ \\
P$\downarrow$ & $2.31 \pm 0.23$ & $2.08 \pm 0.38$ & $0.80 \pm 0.35$ & $1.98 \pm 0.44$ & $5.53 \pm 0.13$ & $5.53 \pm 0.13$ & $4.28 \pm 0.40$ & $4.58 \pm 0.30$ \\ \midrule
\multicolumn{9}{c}{\textbf{Concept: Door}} \\ \midrule
R$\uparrow$ & $-0.50 \pm 1.20$ & $-2.49 \pm 2.59$ & $3.06 \pm 2.30$ & $2.89 \pm 1.97$ & $7.75 \pm 2.39$ & $11.93 \pm 2.70$ & $12.05 \pm 4.10$ & $11.46 \pm 3.55$ \\
S$\downarrow$ & $0.39 \pm 0.02$ & $0.33 \pm 0.02$ & $0.37 \pm 0.02$ & $0.36 \pm 0.02$ & $0.34 \pm 0.02$ & $0.36 \pm 0.01$ & $0.35 \pm 0.03$ & $0.34 \pm 0.03$ \\
P$\downarrow$ & $0.21 \pm 0.22$ & $0.92 \pm 0.39$ & $0.70 \pm 0.33$ & $0.61 \pm 0.32$ & $2.24 \pm 0.31$ & $3.25 \pm 0.35$ & $3.73 \pm 0.34$ & $3.69 \pm 0.33$ \\ \midrule
\multicolumn{9}{c}{\textbf{Concept: Guitar}} \\ \midrule
R$\uparrow$ & $1.23 \pm 1.00$ & $3.02 \pm 2.16$ & $2.14 \pm 1.63$ & $8.84 \pm 1.63$ & $10.06 \pm 0.95$ & $11.49 \pm 1.30$ & $4.49 \pm 2.36$ & $4.49 \pm 2.36$ \\
S$\downarrow$ & $0.35 \pm 0.02$ & $0.37 \pm 0.02$ & $0.34 \pm 0.03$ & $0.34 \pm 0.01$ & $0.34 \pm 0.00$ & $0.33 \pm 0.01$ & $0.36 \pm 0.04$ & $0.36 \pm 0.04$ \\
P$\downarrow$ & $1.11 \pm 0.37$ & $3.28 \pm 0.32$ & $1.86 \pm 0.32$ & $4.62 \pm 0.35$ & $5.72 \pm 0.27$ & $5.06 \pm 0.42$ & $5.35 \pm 0.96$ & $5.35 \pm 0.96$ \\ \midrule
\multicolumn{9}{c}{\textbf{Concept: Ocean}} \\ \midrule
R$\uparrow$ & $5.68 \pm 1.06$ & $13.96 \pm 1.41$ & $1.29 \pm 0.88$ & $4.48 \pm 1.90$ & $-6.12 \pm 1.74$ & $6.81 \pm 1.30$ & $11.69 \pm 2.79$ & $15.14 \pm 1.66$ \\
S$\downarrow$ & $0.32 \pm 0.02$ & $0.31 \pm 0.01$ & $0.32 \pm 0.01$ & $0.32 \pm 0.02$ & $0.45 \pm 0.02$ & $0.35 \pm 0.01$ & $0.32 \pm 0.03$ & $0.35 \pm 0.02$ \\
P$\downarrow$ & $1.31 \pm 0.19$ & $2.15 \pm 0.14$ & $0.24 \pm 0.13$ & $1.12 \pm 0.11$ & $4.68 \pm 0.12$ & $2.74 \pm 0.11$ & $4.31 \pm 1.04$ & $5.10 \pm 0.75$ \\ \midrule
\multicolumn{9}{c}{\textbf{Concept: Rain}} \\ \midrule
R$\uparrow$ & $10.46 \pm 1.94$ & $17.19 \pm 1.46$ & $4.88 \pm 1.83$ & $15.93 \pm 1.75$ & $10.06 \pm 3.10$ & $9.39 \pm 2.23$ & $9.32 \pm 4.24$ & $13.01 \pm 3.53$ \\
S$\downarrow$ & $0.26 \pm 0.02$ & $0.29 \pm 0.02$ & $0.28 \pm 0.01$ & $0.31 \pm 0.02$ & $0.33 \pm 0.03$ & $0.36 \pm 0.03$ & $0.37 \pm 0.06$ & $0.41 \pm 0.04$ \\
P$\downarrow$ & $2.21 \pm 0.41$ & $3.72 \pm 0.24$ & $1.17 \pm 0.72$ & $2.98 \pm 0.67$ & $3.43 \pm 0.54$ & $4.26 \pm 0.49$ & $4.07 \pm 0.71$ & $4.42 \pm 0.57$ \\ \midrule
\multicolumn{9}{c}{\textbf{Overall Score (Averaged across concepts)}} \\  \midrule
R$\uparrow$ & $2.95 \pm 1.57$ & $5.78 \pm 1.42$ & $2.31 \pm 1.60$ & $6.04 \pm 2.04$ & $4.21 \pm 1.79$ & $6.13 \pm 1.72$ & $5.56 \pm 2.24$ & $6.75 \pm 2.23$ \\
S$\downarrow$ & $0.35 \pm 0.02$ & $0.35 \pm 0.02$ & $0.35 \pm 0.02$ & $0.34 \pm 0.02$ & $0.39 \pm 0.02$ & $0.39 \pm 0.02$ & $0.38 \pm 0.03$ & $0.39 \pm 0.03$ \\
P$\downarrow$ & $1.50 \pm 0.37$ & $2.73 \pm 0.32$ & $0.96 \pm 0.43$ & $2.13 \pm 0.42$ & $4.12 \pm 0.29$ & $4.21 \pm 0.32$ & $3.76 \pm 0.71$ & $4.19 \pm 0.59$ \\
\bottomrule
\end{tabular}}
\end{table*}

\end{document}

%% file: math_commands.tex
\usepackage{amsmath,amsfonts,bm}

\def\eqref#1{equation~\ref{#1}}

\def\1{\bm{1}}

\DeclareMathAlphabet{\mathsfit}{\encodingdefault}{\sfdefault}{m}{sl}
\SetMathAlphabet{\mathsfit}{bold}{\encodingdefault}{\sfdefault}{bx}{n}

